\documentclass[11pt]{article}

\usepackage[final]{acl}
\usepackage{times}
\usepackage{latexsym}
\usepackage[T1]{fontenc}
\usepackage[utf8]{inputenc}
\usepackage{microtype}
\usepackage{inconsolata}
\usepackage{graphicx}
\usepackage{amsmath}
\usepackage{amsfonts}
\usepackage{multirow}
\usepackage{booktabs}
\usepackage{setspace}
\usepackage{enumitem}
\usepackage{cleveref}
\usepackage{subcaption}
\usepackage{graphicx}
\usepackage{algorithm}
\usepackage{algpseudocode}
\usepackage[utf8]{inputenc}
\usepackage{colortbl,xcolor}
\usepackage{tabularx}
\usepackage{enumitem}
\usepackage{url}
\usepackage{cuted}
\usepackage{stfloats}

\crefformat{section}{§#2#1#3}

\title{From $\boldsymbol{\log\pi}$ to $\boldsymbol{\pi}$: Taming Divergence in Soft Clipping via Bilateral Decoupled Decay of Probability Gradient Weight}

\author{
    Xiaoliang Fu\textsuperscript{1,2,*}, 
    Jiaye Lin\textsuperscript{1,*,$\dagger$}, 
    Yangyi Fang\textsuperscript{1,3,*}, 
    Chaowen Hu\textsuperscript{1}, \\
    \textbf{Cong Qin}\textsuperscript{1,4}, 
    \textbf{Zekai Shao}\textsuperscript{2}, 
    \textbf{Binbin Zheng}\textsuperscript{1,5}, 
    \textbf{Lu Pan}\textsuperscript{1}, 
    \textbf{Ke Zeng}\textsuperscript{1} \\
    \textsuperscript{1}Meituan \quad
    \textsuperscript{2}Fudan University \quad
    \textsuperscript{3}Tsinghua University \\
    \textsuperscript{4}Peking University \quad
    \textsuperscript{5}University of Science and Technology of China \\
    \texttt{\{fuxiaoliang04, linjiaye\}@meituan.com}
}

\definecolor{mistyblue}{RGB}{230, 237, 245}
\definecolor{almond}{RGB}{250, 241, 230}
\definecolor{ourshighlight}{gray}{0.95}
\definecolor{biascolor}{RGB}{200, 0, 0}
\definecolor{bestcolor}{RGB}{55, 85, 140}
\definecolor{secondcolor}{RGB}{0, 100, 80}
\definecolor{biascolor}{RGB}{192, 0, 0}
\definecolor{improvecolor}{RGB}{45, 60, 145}

\newcommand{\best}[1]{\textbf{#1}}
\newcommand{\second}[1]{\underline{#1}}
\newcommand{\improve}[1]{\textcolor{improvecolor}{#1}}

\newcommand{\modelheader}[2]{\multicolumn{15}{c}{\cellcolor{#1}\textbf{\textsc{#2}}}}
\begin{document}

\maketitle

\begingroup
  \renewcommand\thefootnote{}
  \footnotetext{
    \textsuperscript{*}\ Equal contribution. \quad 
    \textsuperscript{$\dagger$}\ Corresponding author.
  }
\endgroup

\begin{abstract}
Reinforcement Learning with Verifiable Rewards (RLVR) has catalyzed a leap in Large Language Model (LLM) reasoning, yet its optimization dynamics remain fragile. Standard algorithms like GRPO enforce stability via ``hard clipping'', which inadvertently stifles exploration by discarding gradients of tokens outside the trust region. While recent ``soft clipping'' methods attempt to recover these gradients, they suffer from a critical challenge: relying on \textit{log-probability gradient} ($\nabla_\theta\log \pi_\theta$) yields divergent weights as probabilities vanish, destabilizing LLM training. We rethink this convention by establishing \textit{probability gradient} ($\nabla_\theta \pi_\theta$) as the superior optimization primitive. Accordingly, we propose \textbf{D}ecoupled \textbf{G}radient \textbf{P}olicy \textbf{O}ptimization (DGPO), which employs a decoupled decay mechanism based on importance sampling ratios. By applying asymmetric, continuous decay to boundary tokens, DGPO resolves the conflict between stability and sustained exploration. Extensive experiments across DeepSeek-R1-Distill-Qwen series models (1.5B/7B/14B) demonstrate that DGPO consistently outperforms strong baselines on various mathematical benchmarks, offering a robust and scalable solution for RLVR. Our code and implementation are available at: \href{https://github.com/FlyTune/DGPO-RL}{https://github.com/FlyTune/DGPO-RL}.
\end{abstract}

\section{Introduction}
Reinforcement Learning (RL) has emerged as a transformative paradigm for aligning Large Language Models (LLMs) with human intent, shifting focus from imitation learning to goal-driven optimization~\cite{ouyang2022training, shao2024deepseekmath, guo2025deepseek}. In reasoning-intensive domains like mathematics, Reinforcement Learning with Verifiable Rewards (RLVR) achieves remarkable success by leveraging ground-truth feedback to enhance accuracy and logical coherence~\cite{lightman2023let, shao2024deepseekmath}. Beyond standard benchmarks, stronger reasoning has also shown promise in optimization modeling, tool-augmented mathematical problem solving, scientific explanation, and narrative generation~\cite{liu2026automated, luo2025agentmath, shao2025unlocking, shao2025narrative}.
\begin{figure}
\centering
\includegraphics[width=1.0\linewidth]{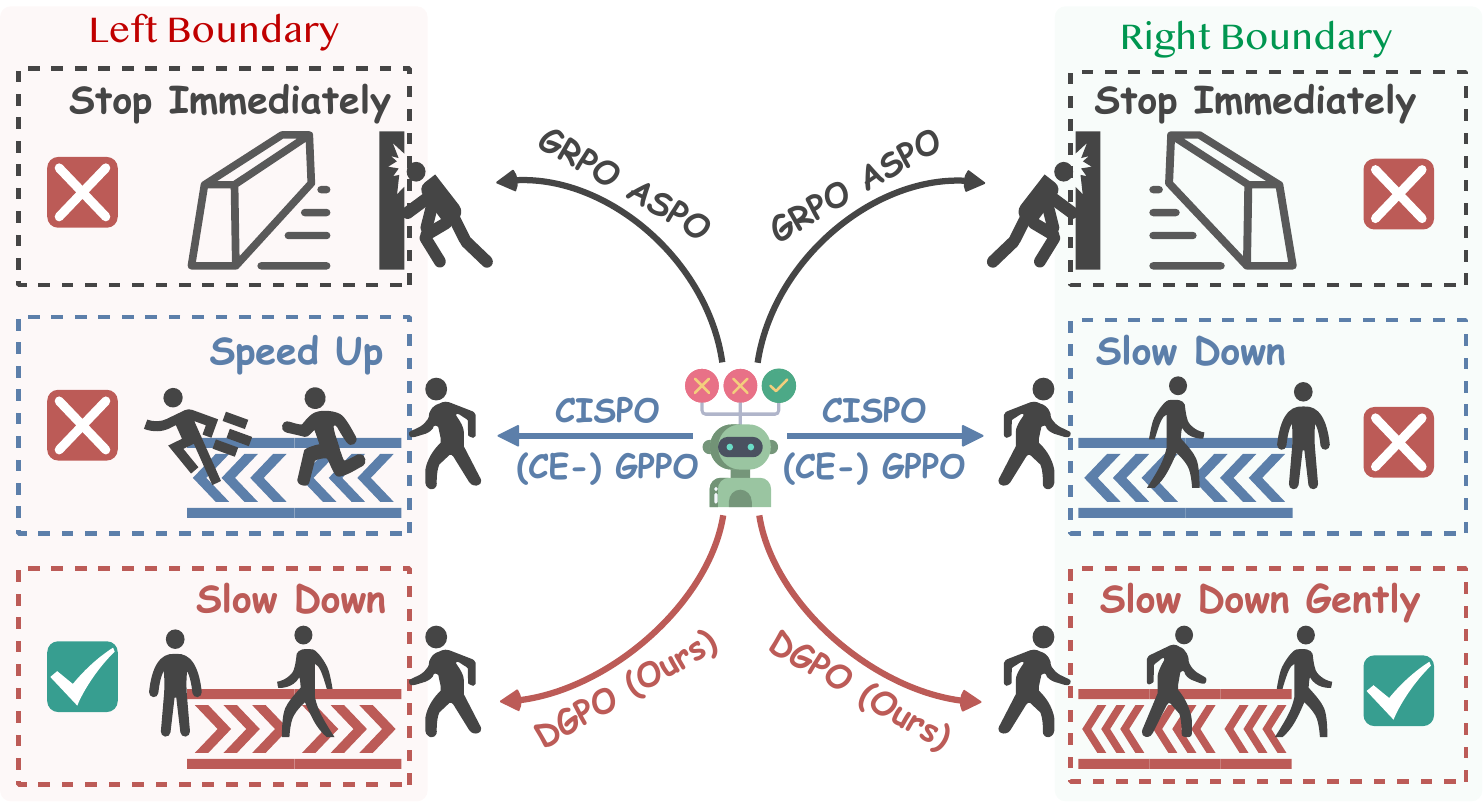}
\caption{Schematic overview of our DGPO algorithm. While ``hard clipping'' methods (GRPO/ASPO) discard gradients at boundaries, and prior ``soft clipping'' approaches (CISPO/GPPO/CE-GPPO) risk divergence on the left boundary and limit exploration on the right boundary, DGPO optimizes the gradient decay mechanism accordingly. By enforcing a controlled ``Slow Down'' behavior for stability and a ``Slow Down Gently'' behavior to sustain exploration, DGPO effectively resolves the exploration-stability conflict while maintaining minimal bias against the true policy gradient.}
\vspace{-10pt}
\label{fig:intro}
\end{figure}

Despite this promise, RLVR optimization remains fragile from the conflict between exploration and stability. Algorithms like PPO~\cite{schulman2015trust} and GRPO~\cite{shao2024deepseekmath} enforce a trust region by ``hard clipping'' the Importance Sampling (IS) ratio ($\pi_\theta / \pi_{\theta_{\text{old}}}$). While preventing destructive policy shifts, this inadvertently induces a \textit{vanishing gradient for exploration}. Tokens drifting outside the trust region -- representing valuable exploratory steps -- receive zero updates, leading to entropy collapse and premature convergence~\cite{williams1991function, eysenbach2021maximum}.

Recently, ``soft clipping'' approaches attempt to preserve gradients for out-of-bound tokens~\cite{chen2025minimax, su2025klear, su2025gppo}. However, we identify a fundamental challenge: these methods predominantly operate on \textit{log-probability gradient} ($\nabla_\theta \log \pi_\theta$). As probability approaches zero, the gradient weight of log-probability-based methods grows divergently. This causes catastrophic instability at the left boundary (where the IS ratio is extremely low), disproportionately penalizing tokens with vanishing probabilities.

Accordingly, we propose a paradigm shift: establishing \textit{probability gradient} ($\nabla_\theta \pi_\theta$) as the primary optimization primitive over \textit{log-probability gradient} ($\nabla_\theta \log \pi_\theta$). By analyzing the gradient landscape in probability space, we derive criteria for optimal weighting: (1) sustain exploration for clipped tokens, (2) ensure stability via convergent weights at boundaries, and (3) maximize alignment with the unbiased policy gradient.

In this paper, we propose \textbf{D}ecoupled \textbf{G}radient \textbf{P}olicy \textbf{O}ptimization (DGPO) guided by these principles. DGPO replaces ``hard clipping'' with a decoupled decay mechanism applied to the probability gradient weight. It applies a polynomial decay to tokens with \textit{low IS ratios} (left boundary) for stability, and a reciprocal radical decay to tokens with \textit{high IS ratios} (right boundary) to foster exploration. This mathematically guarantees gradient continuity and prevents the weight divergence seen in prior methods. Our contributions are summarized as:
\begin{itemize}[leftmargin=0.5cm]
\item We introduce a novel perspective, establishing the gradient of probability, rather than log-probability, as the superior optimization primitive in LLMs, with two key insights: the inherent alignment of RL objectives and the geometric symmetry of probability space, which facilitates stable gradient design.
\item We propose DGPO, which leverages a decoupled adaptive decay mechanism to reconcile the exploration-stability conflict.  Crucially, this design preserves gradients for clipped tokens while rigorously preventing weight divergence.
\item Comprehensive experiments against competitive baselines across mathematical reasoning benchmarks demonstrate the effectiveness of DGPO. Further results on diverse models scales confirm its scalability and robustness.
\end{itemize}

\section{Related Works}

\subsection{RLVR in LLMs}
RLVR utilizes deterministic signals (e.g., correct answers) rather than learned reward models to enhance LLM reasoning~\cite{uesato2022solving}. Recent advancements like DeepSeek-Math~\cite{shao2024deepseekmath} popularized GRPO, which efficiently normalizes rewards within a sampled group, eliminating the need for a critic. However, GRPO inherits the clipping-induced exploration limitations of PPO.

\subsection{Importance Sampling and Clipping}
IS enables off-policy training by correcting distribution shifts via the ratio $\pi_\theta / \pi_{\theta_{\text{old}}}$~\cite{precup2000eligibility, schulman2015trust}. To prevent variance explosion, PPO~\cite{schulman2017proximal} clips this ratio within $[1-\varepsilon, 1+\varepsilon]$. This ``hard clipping'' mechanism zeros out gradients for outlier tokens, prioritizing exploitation while neglecting low-probability tokens essential for exploration, causing rapid entropy decay~\cite{o2016combining, yu2025dapo, bai2026ttvs, zhou2026look}. To mitigate exploration losses, several methods dynamically adjust clipping bounds~\cite{yu2025dapo, yang2025dcpo}, allowing more updates for specific tokens. However, they still rely on hard boundaries, inevitably discarding gradient information for tokens beyond the adjusted thresholds.

\subsection{Soft Clipping Policy Optimization} 
More recent approaches replace ``hard clipping'' with soft schemes. For instance, CISPO~\cite{chen2025minimax} combines ``soft clipping'' with soft dual clip~\cite{ye2020mastering}, while GPPO~\cite{su2025klear} retains a constant log-probability gradient weight for out-of-bound tokens. Crucially, both CISPO and GPPO suffer from left boundary instability: as $\pi_\theta \to 0$, the gradient grows indefinitely. Without proper decay mechanisms, this results in divergent updates that destabilize training. CE-GPPO~\cite{su2025gppo} attempts to scale boundary gradients via hyperparameters but fails to resolve the underlying divergence. Distinctly, ASPO~\cite{wang2025aspo} proposes a reversed ratio to balance updates. Compared with these approaches, DGPO redefines the optimization target in probability space to ensure theoretical continuity and stability.

\section{Preliminary}
\paragraph{Problem Definition.}
In this paper, we focus on RLVR settings. Given a query $q$ from dataset $\mathcal{D}$, a policy $\pi_\theta$ generates a response $o$. The rule-based reward function $r(q,o) \in \{-1,1\}$ evaluates the correctness of each response. Following GRPO~\cite{shao2024deepseekmath}, for each query $q$, a group of $G$ outputs $\{o_i\}_{i=1}^G$ is sampled from the old policy $\pi_{\theta_{\text{old}}}$. The advantage $\hat{A}_i$ is computed by normalizing the group-level rewards: $\hat{A}_i = (r(q, o_i) - \mu_R) / \sigma_R$, where $\mu_R$ and $\sigma_R$ denote the mean and standard deviation of the rewards within the group.

\paragraph{Unified Gradient Formulation.}
To align off-policy training with reward maximization while constraining policy shifts for stability, most RLVR algorithms employ IS and clipping mechanisms. Thus, we formulate a \textit{unified gradient estimator} that encompasses these methods. Let $w_{i,t}(\theta) = \frac{\pi_\theta(o_{i,t}|q, o_{i,<t})}{\pi_{\theta_{\text{old}}}(o_{i,t}|q, o_{i,<t})}$ denotes the token-level IS ratio, and the gradient is then expressed as:
\begin{flalign}
&\nabla_\theta\mathcal{J}(\theta) = \mathbb{E}_{q \sim \mathcal{D}, \{o_i\}_{i=1}^G \sim \pi_{\theta_{\text{old}}}(\cdot|q)} \frac{1}{\sum_{i=1}^G |o_i|} \nonumber \\
&\sum_{i=1}^G \sum_{t=1}^{|o_i|} \mathcal{F}_{i,t}(\theta) \hat{A}_{i} \nabla_\theta \log \pi_\theta(o_{i,t}|q,o_{i,<t}).
\end{flalign}
The behavior of $\mathcal{F}_{i,t}(\theta)$ typically depends on five distinct regions, which are simultaneously defined by the clipping boundaries ($1-\varepsilon_\text{low}$ and $1+\varepsilon_\text{high}$) and the sign of the advantage $\hat{A}_i$. We formally define these regions in Table~\ref{tab:case_definition}.

\begin{table}[htbp]
\centering
\renewcommand{\arraystretch}{1.2}
\setlength{\tabcolsep}{4.8pt}
\small
\caption{Definition of clipping regions based on IS ratio $w_{i,t}(\theta)$ and advantage $\hat{A}_i$. Abbrev. denotes the abbreviation of five cases under different conditions.}
\begin{tabular}{l c c}
\toprule
\textbf{Case} & \textbf{Abbrev.} & \textbf{Condition} \\
\midrule
Left Boundary  & \multirow{2}{*}{LN} &
$w_{i,t}(\theta) < 1 - \varepsilon_{\text{low}}$ \\
\quad (Low ratio, Neg. adv.) & & $\land \ \hat{A}_{i} < 0$ \\
\addlinespace[2pt]
Right Boundary & \multirow{2}{*}{HP} & 
$w_{i,t}(\theta) > 1 + \varepsilon_{\text{high}}$ \\
\quad (High ratio, Pos. adv.) & & $\land \ \hat{A}_{i} > 0$ \\
\addlinespace[2pt]
Reverse Left Boundary & \multirow{2}{*}{LP} & 
$w_{i,t}(\theta) < 1 - \varepsilon_{\text{low}}$ \\
\quad (Low ratio, Pos. adv.) & & $\land \ \hat{A}_{i} > 0$ \\
\addlinespace[2pt]
Reverse Right Boundary & \multirow{2}{*}{HN} & 
$w_{i,t}(\theta) > 1 + \varepsilon_{\text{high}}$ \\
\quad (High ratio, Neg. adv.) & & $\land \ \hat{A}_{i} < 0$ \\
\addlinespace[2pt]
In-Boundary & \multirow{2}{*}{M} & 
$\neg\text{LN} \land \neg\text{HP}$ \\
\quad (Medium ratio) & & $\land \neg\text{LP} \land \neg\text{HN}$ \\
\bottomrule
\end{tabular}
\label{tab:case_definition}
\end{table}

\paragraph{Instantiations of Clipping Strategies.}
Various existing methods can be interpreted as specific instantiations of $\mathcal{F}_{i,t}(\theta)$. Standard \textbf{PPO}~\cite{schulman2017proximal} and \textbf{GRPO}~\cite{shao2024deepseekmath} apply ``hard clipping'' to penalize excessive updates: 
\begin{flalign}
\mathcal{F}^\text{GRPO}_{i,t}(\theta) = 
\begin{cases} 
0, & \text{if } \text{LN} \lor \text{HP}, \\
w_{i,t}(\theta), & \text{otherwise}.
\end{cases}
\end{flalign}
\textbf{CISPO}~\cite{chen2025minimax} preserves exploration for these tokens by the following gradient weight:
\begin{flalign}
\mathcal{F}^\text{CISPO}_{i,t}(\theta) = 
\begin{cases} 
1 - \varepsilon_\text{low}, & \text{if } \text{LN} \lor \text{LP}, \\
1 + \varepsilon_\text{high}, & \text{if } \text{HP} \lor \text{HN}, \\
w_{i,t}(\theta), & \text{otherwise}.
\end{cases}
\end{flalign}
This method inherently employs soft dual clip ~\cite{ye2020mastering}, which clips the loss values for LP and HN (reverse cases) while preserving gradients. Subsequently, \textbf{GPPO} ~\cite{su2025klear} introduces a gradient weight more aligned with PPO, focusing solely on LN and HP cases:
\begin{flalign}
\mathcal{F}^\text{GPPO}_{i,t}(\theta) =
\begin{cases}
1 - \varepsilon_\text{low}, & \text{if } \text{LN}, \\
1 + \varepsilon_\text{high}, & \text{if } \text{HP}, \\
w_{i,t}(\theta), & \text{otherwise}.
\end{cases}
\end{flalign}
\textbf{CE-GPPO} ~\cite{su2025gppo} further refines GPPO by introducing hyperparameters ($\beta_1$ and $\beta_2$) to control gradient scaling at both boundaries:
\begin{flalign}
\mathcal{F}^\text{CE}_{i,t}(\theta) =
\begin{cases} 
\beta_1 (1 - \varepsilon_\text{low}), & \text{if } \text{LN}, \\
\beta_2 (1 + \varepsilon_\text{high}), & \text{if } \text{HP}, \\
w_{i,t}(\theta), & \text{otherwise}.
\end{cases}
\end{flalign}
To address the imbalance in updating positive-advantage tokens, \textbf{ASPO} ~\cite{wang2025aspo} reverses gradient weight and adopts soft dual clip:
\begin{flalign}
\mathcal{F}^\text{ASPO}_{i,t}(\theta) = 
\begin{cases} 
0, & \text{if } \text{LN} \lor \text{HP}, \\
1 - \varepsilon_\text{low}', & \text{if } \text{LP}', \\
1 + \varepsilon_\text{high}', & \text{if } \text{HN}', \\
\frac{1}{w_{i,t}(\theta)}, & \text{otherwise}.
\end{cases}
\end{flalign}
The values of $1 - \varepsilon_\text{low}'$ and $1 + \varepsilon_\text{high}'$ for cases LP$'$ and HN$'$ in ASPO may be more extreme.

\section{Methodology}

\begin{figure*}[t]
\centering
\includegraphics[width=1.0\linewidth]{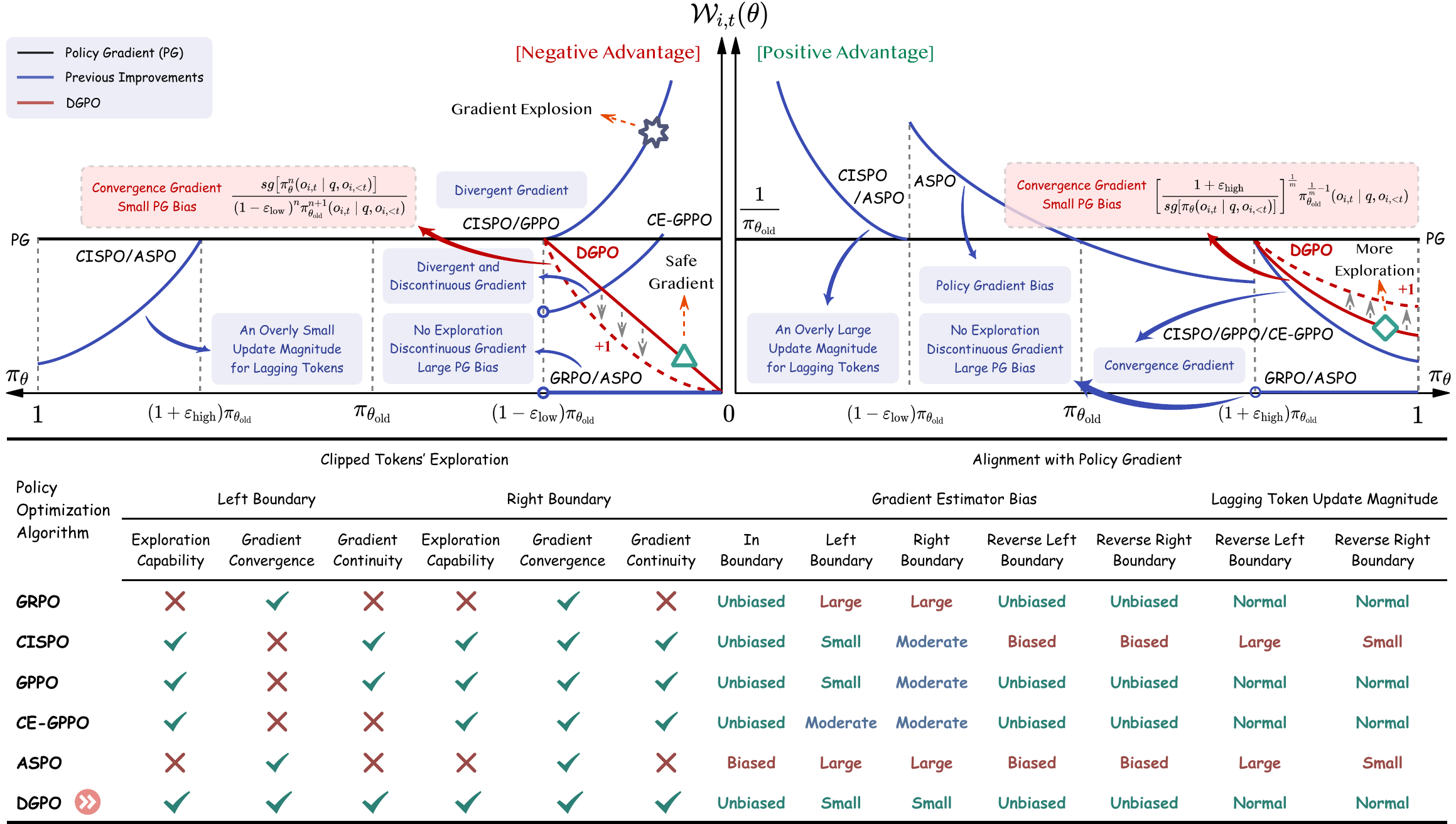}
\caption{Comparative analysis of gradient dynamics. We systematically contrast DGPO with the standard GRPO, prior ``soft clipping'' enhancements (CISPO, GPPO, and CE-GPPO), and importance sampling improvements (ASPO). The visualization highlights the theoretical properties regarding the exploration capability of clipped tokens and the alignment with the true policy gradient, demonstrating DGPO's superior stability and gradient consistency.}
\vspace{-1em}
\label{fig:method}
\end{figure*}

\subsection{Shifting Focus: From $\boldsymbol{\log\pi}$ to $\boldsymbol{\pi}$}
Classical policy gradient methods~\cite{williams1992simple} typically employ the log-derivative trick to reformulate the gradient into an expectation, which establishes $\nabla_\theta\log\pi_\theta$ as the canonical term. This has led to a pervasive focus on log-probability gradient in subsequent research. However, we rethink that \textit{probability gradient} constitutes a superior analytical and design target for two primary reasons: (1) probability, rather than log-probability, acts as the superior optimization primitive in LLM training, and (2) probability exhibits superior geometric symmetry within its value range.

\paragraph{Probability as the Optimization Primitive.}
To substantiate this claim, we contrast the token-level objectives of Supervised Fine-Tuning (SFT) and RL. SFT maximizes the mean of expert token log-probabilities, with the gradient estimator as:
\begin{flalign}
\small
\label{equ:sft_estimator}
\nabla_\theta\mathcal{J}_{\text{SFT}}(\theta) &= \mathbb{E}_{\mathcal{D}} \sum_{t=1}^{|o|} \nabla_\theta \log\pi_\theta(o_t|q, o_{<t}).
\end{flalign}
Conversely, RL (under binary advantage assumptions) equivalently maximizes the mean of expert token \textit{probabilities} (derivation provided in Appendix \ref{app:rl_sft_form_derivation}), yielding the estimator:
\begin{flalign}
\small
\label{equ:rl_estimator}
\nabla_\theta\mathcal{J}_{\text{RL}}(\theta) = \mathbb{E}_{\mathcal{D}} \sum_{t=1}^{|o|} \nabla_\theta \pi_\theta(o_t \mid q, o_{<t}).
\end{flalign}
Equations~\eqref{equ:sft_estimator} and~\eqref{equ:rl_estimator} reveal a fundamental distinction: SFT operates on log-probability, whereas RL inherently operates on probability. This distinction renders the SFT objective mathematically a lower bound of the RL objective~\cite{qin2025supervised}, explaining the empirical performance ceiling often observed in SFT compared with RL~\cite{wu2025generalization}. Consequently, we identify probability as the superior optimization primitive. Since the probability aligns more closely with the core requirements of LLM training, its gradient should be prioritized in algorithm design.

\paragraph{Geometric Symmetry and Boundedness.} 
In LLM training, token probabilities reside in the symmetric and bounded interval $(0, 1)$. This boundedness facilitates the analysis of gradient impacts on probability values and enables the design of symmetric gradient mechanisms. In contrast, log-probabilities span the asymmetric and unbounded interval $(-\infty, 0)$, complicating the gradient design.

\begin{table*}[htbp]
\centering
\setlength{\tabcolsep}{8.7pt}
\footnotesize
\caption{Comparative analysis of gradient bias. We evaluate the bias magnitudes of DGPO and other baselines relative to the true policy gradient under different boundary conditions.}
\begin{tabular}{l c}
\toprule
\textbf{Bias Condition} & \textbf{Magnitude Comparison}\\
\midrule
In-Boundary Bias &
$0 = \textcolor{biascolor}{\text{Bias}_\text{DGPO}^\text{M}} = \text{Bias}_\text{GRPO}^\text{M} = \text{Bias}_\text{CISPO}^\text{M} = \text{Bias}_\text{GPPO}^\text{M} = \text{Bias}_\text{CE}^\text{M} \leq \text{Bias}_\text{ASPO}^\text{M}$ \\
\addlinespace[2pt]
Left-Boundary Bias &
$\begin{cases}
    0 < \textcolor{biascolor}{\text{Bias}_\text{DGPO}^\text{LN}} < \text{Bias}_\text{CISPO}^\text{LN} = \text{Bias}_\text{GPPO}^\text{LN} < \text{Bias}_\text{CE}^\text{LN} < \text{Bias}_\text{GRPO}^\text{LN} = \text{Bias}_\text{ASPO}^\text{LN}, & n=1 \\
    0 < \text{Bias}_\text{CISPO}^\text{LN} = \text{Bias}_\text{GPPO}^\text{LN} < \textcolor{biascolor}{\text{Bias}_\text{DGPO}^\text{LN}} < \text{Bias}_\text{CE}^\text{LN} < \text{Bias}_\text{GRPO}^\text{LN} = \text{Bias}_\text{ASPO}^\text{LN}, & n>1
\end{cases}$ \\
\addlinespace[2pt]
Right-Boundary Bias &
$0 < \textcolor{biascolor}{\text{Bias}_\text{DGPO}^\text{HP}} \leq \text{Bias}_\text{CISPO}^\text{HP} = \text{Bias}_\text{GPPO}^\text{HP} = \text{Bias}_\text{CE}^\text{HP} < \text{Bias}_\text{GRPO}^\text{HP} = \text{Bias}_\text{ASPO}^\text{HP}$ \\
\addlinespace[2pt]
Reverse Left-Boundary Bias &
$0 = \textcolor{biascolor}{\text{Bias}_\text{DGPO}^\text{LP}} = \text{Bias}_\text{GRPO}^\text{LP} = \text{Bias}_\text{GPPO}^\text{LP} = \text{Bias}_\text{CE}^\text{LP} < \text{Bias}_\text{CISPO}^\text{LP} = \text{Bias}_\text{ASPO}^\text{LP}$ \\
\addlinespace[2pt]
Reverse Right-Boundary Bias &
$0 = \textcolor{biascolor}{\text{Bias}_\text{DGPO}^\text{HN}} = \text{Bias}_\text{GRPO}^\text{HN} = \text{Bias}_\text{GPPO}^\text{HN} = \text{Bias}_\text{CE}^\text{HN} < \text{Bias}_\text{CISPO}^\text{HN} = \text{Bias}_\text{ASPO}^\text{HN}$ \\
\bottomrule
\end{tabular}
\label{tab:bias_comparison}
\vspace{-1em}
\end{table*}

\subsection{Decoupled Gradient Policy Optimization}

\paragraph{Instability in Soft Clipping.}
While clipped tokens often contain critical information essential for model performance~\cite{liu2025uniform, su2025klear}, preserving their gradients requires a principled approach. Prior ``soft clipping'' works~\cite{chen2025minimax, su2025klear} maintain constant log-probability gradient weights in LP and HN cases. However, the gradient weights are present asymmetrically at the boundaries, contradicting the symmetric nature of probability values. 

Thus, we define probability gradient weight as $\mathcal{W}_{i,t}(\theta) = \frac{\nabla_{\theta} \mathcal{J}(\theta)}{\hat{A}_{i} \nabla_{\theta} \pi_\theta(o_{i,t} | q, o_{i,<t})}$. As illustrated in Figure \ref{fig:method}, while the right-boundary gradient weight decreases convergently (promoting stability), the left-boundary weight grows divergently. This causes negative-advantage tokens to shrink disproportionately—the smaller the probability, the larger the update magnitude—leading to training instability~\cite{su2025gppo, zhai2026maximizing, zhang2026logicalphasetransitionsunderstanding}. Although CE-GPPO attempts to mitigate this via hyperparameters, it does not resolve the divergent growth, leaving the risk of collapse.

\paragraph{DGPO Formulation.}
To address these limitations, we propose DGPO in this paper, designed to: (1) preserve gradient for clipped tokens, (2) stabilize exploration via adaptive gradient decay, and (3) minimize policy gradient bias for alignment.

We define the weighting function $\mathcal{W}^\text{DGPO}_{i,t}(\theta)$ based on the regions defined in Table \ref{tab:case_definition}:
\begin{equation}
\label{eq:dgpo_weight}
\small
\mathcal{W}^\text{DGPO}_{i,t}(\theta) = 
\begin{cases} 
    C_{\text{left}} \cdot sg^n[\pi_\theta(o_{i,t}|q,o_{i,<t})], & \text{if LN}, \\
    C_{\text{right}} \cdot sg^{-\frac{1}{m}}[\pi_\theta(o_{i,t}|q,o_{i,<t})], & \text{if HP}, \\
    \frac{1}{\pi_{\theta_{\text{old}}}}, & \text{otherwise}.
\end{cases}
\end{equation}
Then, the objective function is formulated as:
\begin{flalign}
\label{equ:dgpo_loss}
&\mathcal{J}_{\text{DGPO}}(\theta) = \mathbb{E} _{q \sim \mathcal{D}, \{o_i\}_{i=1}^G \sim \pi_{\theta_{\text{old}}}(\cdot|q)} \frac{1}{\sum_{i=1}^G |o_i|} \nonumber\\ 
&\sum_{i=1}^G \sum_{t=1}^{|o_i|} \mathcal{W}^\text{DGPO}_{i,t}(\theta) \hat{A}_{i} \pi_\theta(o_{i,t}|q,o_{i,<t}),
\end{flalign}
where $n, m \in \mathbb{Z}^+$ are hyperparameters controlling the decay rate, and $sg[\cdot]$ denotes the stop-gradient operator. $C_{\text{left}}=(1-\varepsilon_\text{low})^{-n} \pi^{-(n+1)}_{\theta_{\text{old}}}(o_{i,t}|q, o_{i,<t})$ and $C_{\text{right}}=(1+\varepsilon_\text{high})^{\frac{1}{m}}\pi^{\frac{1}{m}-1}_{\theta_\text{old}}(o_{i,t}|q, o_{i,<t})$ are constants ensuring continuity of gradient weights (derivation provided in Appendix \ref{app:continuity_proof}).

\subsection{Theoretical Analysis and Advantages}

\paragraph{Gradient Preserving and Exploration.}
Following prior gradient-preserving methods, DGPO adopts ``soft clipping'' to maintain gradient of clipped tokens. By retaining adaptively reduced gradients in both LN (Low Ratio, Negative Advantage) and HP (High Ratio, Positive Advantage) scenarios, DGPO enables sustained exploration for clipped tokens, enhancing the model's exploration potential and raising its convergence upper bound.

\paragraph{Symmetric Stability Control.}
To ensure training stability while adhering to the principle of trust region optimization (limiting divergence between $\pi_\theta$ and $\pi_{\theta_\text{old}}$), gradient weights must decrease as the policy deviates. DGPO achieves this via a decoupled design: (1) \textbf{Left Boundary:} A positive integer power function of $\pi_\theta$, ensuring weights decay as probability decreases. (2) \textbf{Right Boundary:} A reciprocal radical power function of $\pi_\theta$, ensuring weights decay as probability increases.

Furthermore, DGPO manages entropy dynamics by controlling the ``openness'' of gradient weights. Since LN tokens drive entropy reduction and HP tokens drive entropy increase~\cite{cui2025entropy, su2025gppo}, the default symmetric setting ($m=1$ and $n=1$) may lead to rapid entropy decay. We balance this by adjusting $n$ (reducing left-boundary openness) and $m$ (increasing right-boundary openness). Additionally, multiplying gradient weights by $C_{\text{left}}$ and $C_{\text{right}}$ ensures continuity at the boundaries, ensuring a smooth transition between stable updates and decelerated exploration.

\begin{table*}[t]
\footnotesize
\centering
\renewcommand{\arraystretch}{1.47}
\setlength{\tabcolsep}{3.5pt}
\caption{Comparison results of different methods on various benchmarks. Avg@32 (\%) and Pass@32 (\%) are abbreviated as A@32 and P@32. The best results are \best{bold}, and the second-best results are \second{underlined}, respectively.}
\vspace{-0.5em}
\resizebox{\textwidth}{!}{
\begin{tabular}{lcccccccccccccc}
\toprule
\multirow{2}{*}{\textbf{Method}} & \multicolumn{2}{c}{\textbf{AIME24}} & \multicolumn{2}{c}{\textbf{AIME25}} & \multicolumn{2}{c}{\textbf{AMC23}} & \multicolumn{2}{c}{\textbf{MATH500}} & \multicolumn{2}{c}{\textbf{Minerva}} & \multicolumn{2}{c}{\textbf{Olympiad}} & \multicolumn{2}{c}{\textbf{Avg.}} \\
\cmidrule(lr){2-3} \cmidrule(lr){4-5} \cmidrule(lr){6-7} \cmidrule(lr){8-9} \cmidrule(lr){10-11} \cmidrule(lr){12-13} \cmidrule(lr){14-15}
& A@32 & P@32 & A@32 & P@32 & A@32 & P@32 & A@32 & P@32 & A@32 & P@32 & A@32 & P@32 & A@32 & P@32 \\
\midrule

\modelheader{mistyblue}{DeepSeek-R1-Distill-Qwen-1.5B} \\ 
\midrule
GRPO  & 33.2 & 71.8 &  27.7 & 49.9 & 79.5 & 94.8 & \second{77.6} & 90.8 & 26.1 & 48.8 & \second{46.3} & 64.7 & 48.4 & 70.1 \\
CISPO & 34.8 & 69.1 &  25.8 & 53.3 & 76.9 & \second{94.9} & 76.8 & \best{91.8} & 26.5 & \best{54.2} & 45.8 & \second{65.8} & 47.8 & \second{71.5} \\
GPPO  & 29.6 & 60.5 &  23.5 & 51.9 & 73.5 & 94.1 & 76.3 & 89.1 & 26.6 & 50.0 & 43.9 & 64.2 & 45.6 & 68.3 \\
CE-GPPO & 35.1 & 70.2 &  27.7 & \second{55.1} & 82.5 & \best{95.0} & 76.7 & 90.2 & \second{27.8} & \second{50.5} & 45.6 & 63.1 & \second{49.2} & 70.7 \\
ASPO  & \second{36.4} & \second{73.2} &  \second{28.3} & 51.5 & \second{83.1} & 94.7 & 74.6 & 90.5 & 26.0 & 49.8 & 44.9 & 63.7 & 48.9 & 70.6 \\
\rowcolor{ourshighlight}
\textbf{Ours} & \best{43.3} & \best{79.3} & \best{32.8} & \best{56.1} & \best{86.0} & \best{95.0} & \best{77.9} & \second{91.0} & \best{28.2} & 50.4 & \best{48.0} & \best{66.4} & \best{52.7} & \best{73.0} \\

\midrule

\modelheader{almond}{DeepSeek-R1-Distill-Qwen-7B} \\
\midrule
GRPO  & 48.2 & \best{82.5} & 37.4 & 60.5 & 88.1 & \second{96.6} & \second{84.8} & 92.4 & 37.4 & 57.2 & \second{57.2} & \second{73.9} & 58.9 & \second{77.2} \\
CISPO & 51.6 & 76.6 & \second{38.2} & \second{65.4} & \best{90.6} & \second{96.6} & 82.1 & 91.6 & 38.7 & 56.5 & 54.3 & 69.9 & \second{59.3} & 76.1 \\
GPPO  & 43.1 & 72.5 & 31.7 & 62.5 & 85.6 & 94.9 & 83.2 & \best{95.4} & 33.1 & \best{59.3} & 53.2 & \best{74.3} & 55.0 & 76.5 \\
CE-GPPO & 48.7 & 76.9 & 36.4 & 60.4 & \second{90.5} & 95.0 & 84.3 & 93.3 & \second{39.0} & 55.4 & 54.9 & 72.5 & 59.0 & 75.6 \\
ASPO  & \second{51.8} & 79.6 & 37.1 & 54.1 & 90.0 & \best{97.2} & 83.8 & \second{94.9} & 37.0 & \second{59.2} & 54.1 & 72.8 & 59.0 & 76.3 \\
\rowcolor{ourshighlight}
\textbf{Ours} & \best{55.5} & \second{81.9} & \best{43.1} & \best{68.0} & \best{90.6} & \second{96.6} & \best{85.4} & 92.0 & \best{39.8} & 56.7 & \best{57.7} & 72.0 & \best{62.0} & \best{77.9} \\

\bottomrule
\end{tabular}
}
\vspace{-1em}
\label{tab:main_results}
\end{table*}

\paragraph{Minimal Bias Estimation.}
A critical challenge of existing RL algorithms is their deviation from the distribution correction principle of IS, resulting in significant bias relative to the true policy gradient. DGPO minimizes this bias to establish a stronger theoretical guarantee, which is the key factor for higher convergence bounds of model performance. Table~\ref{tab:bias_comparison} presents a comparative analysis of bias (proofs provided in Appendix~\ref{app:bias_proof}).

While CISPO and GPPO theoretically achieve minimal left-boundary bias when $n>1$, they suffer from divergent gradient weights, constituting the root of training collapse. DGPO balances these objectives: it ensures gradient continuity and adaptive convergence while achieving minimal bias at $n=1$ and maintaining a constant bias ratio relative to CISPO/GPPO for $n>1$, thus offering a trade-off between stability and correctness.

\section{Experiments}
\begin{figure*}[htbp]
\centering
\includegraphics[width=1.0\linewidth]{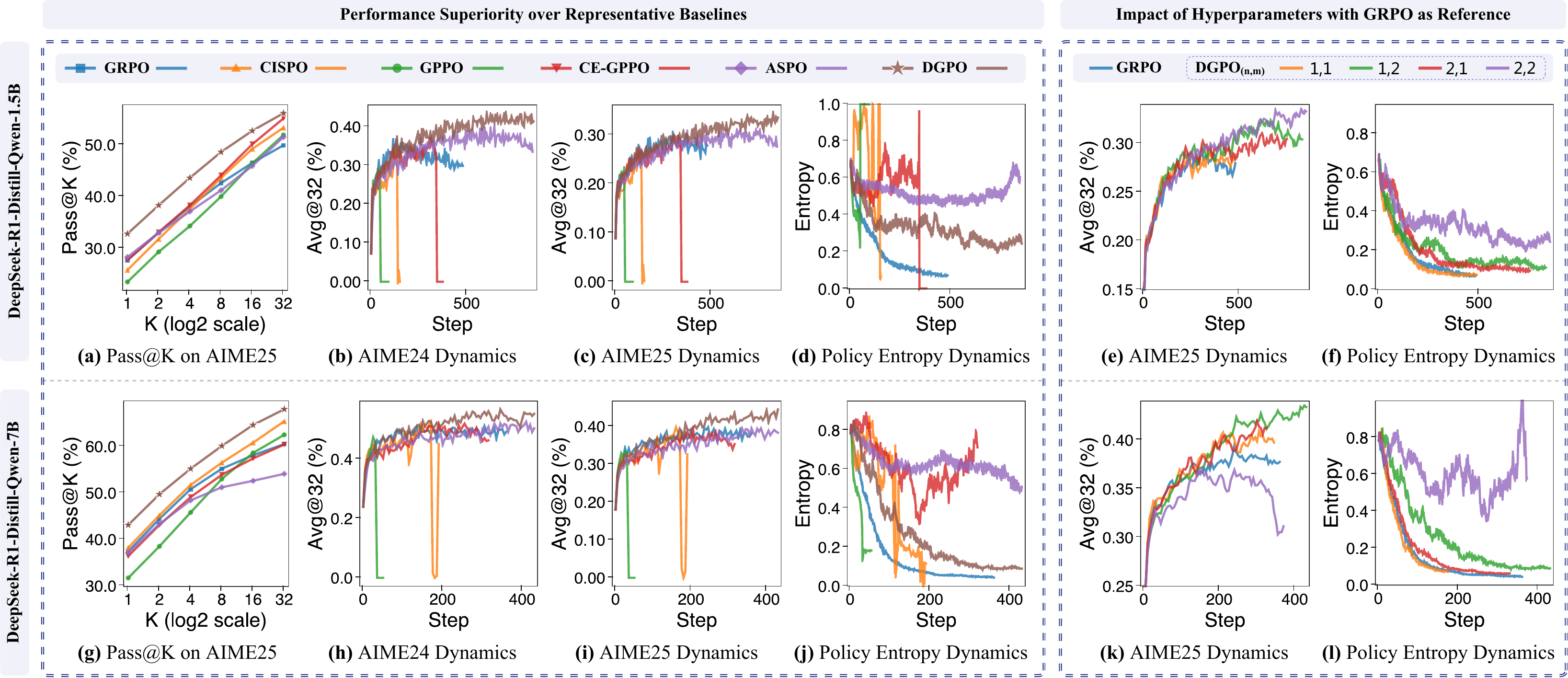}
\caption{Comprehensive comparison of training dynamics and performance. 
\textbf{Top row:} DeepSeek-R1-Distill-Qwen-1.5B results. 
\textbf{Bottom row:} DeepSeek-R1-Distill-Qwen-7B results. 
\textbf{Columns (L-R):} Pass@K on AIME25, Avg@32 on AIME24/25, policy entropy, followed by hyperparameter analysis for Avg@32 and entropy.}
\label{fig:experiment1_final}
\vspace{-8pt}
\end{figure*}

\subsection{Experimental Setup}
\paragraph{Configurations.}
For all experiments, we employ DeepSeek-R1-Distill-Qwen~\cite{guo2025deepseek} with various scales as backbone models to validate the effectiveness of DGPO. The open-source DAPO-Math-17K dataset~\cite{yu2025dapo} is utilized for model training. During training, rule-based rewards are computed with \texttt{math\_verify}~\cite{Kydlicek_Math-Verify_Math_Verification}, while a conservative method \texttt{prime\_math} is used for evaluation~\cite{lightman2023let}. 

Moreover, to ensure consistent convergence dynamics across different model scales, we calibrate the learning rate to maintain a \textit{constant total gradient variance} (derivation provided in Appendix \ref{app:lr_scaling}). Accordingly, the learning rates are set as follows: $1.0 \times 10^{-6}$ for 1.5B, $4.63 \times 10^{-7}$ for 7B, and $3.27 \times 10^{-7}$ for 14B. We set the batch size to 512 and the mini-batch size to 32. This configuration enables 16 off-policy updates per IS step, sufficiently amplifying the influence of boundary tokens. $\varepsilon_\text{low}$ and $\varepsilon_\text{high}$ are fixed to 0.2. Detailed configurations are available in Appendix \ref{app:implementation_details}.

\paragraph{Benchmarks and Metrics.}
We assess the generalization of reasoning capabilities on widely used mathematical benchmarks, i.e., AIME24~\cite{AIME24}, AIME25~\cite{AIME25}, AMC23~\cite{AMC}, MATH500~\cite{hendrycks2021measuring}, Minerva~\cite{lewkowycz2022solving}, and OlympiadBench~\cite{he2024olympiadbench}. More details about benchmarks are described in Appendix \ref{app:implementation_details}. We report Avg@32 to measure the expected performance and Pass@32 to gauge the potential capability.

\paragraph{Baselines.}
We benchmark DGPO against standard GRPO and variants employing distinct gradient weighting strategies, including ``soft clipping'' (CISPO and GPPO), scaled ``soft clipping'' (CE-GPPO, adopting the recommended $\beta_1=0.75$ and $\beta_2=1$), and reverse gradient weight (ASPO).

\subsection{Main Results}
\paragraph{Overall Performance.}
Table~\ref{tab:main_results} presents the performance comparison across 1.5B and 7B scales. DGPO demonstrates superior performance across the majority of benchmarks on both scales. On the 1.5B model, DGPO surpasses the vanilla GRPO by +4.3\% and the best baseline (CE-GPPO) by +3.5\% in average Avg@32. On the 7B model, the improvement is also significant, with DGPO outperforming GRPO by +3.1\% and CISPO by +2.7\%. As shown in Figure \ref{fig:experiment1_final}(a,g), DGPO consistently maintains higher Pass@K scores on AIME2025 compared with all baselines, demonstrating superior potential capability. Detailed results of Pass@K ($\text{K}=1, \dots, 32$) on AIME24 and AIME25 across all model scales are provided in Appendix~\ref{app:detailed_passk}.

\paragraph{Training Dynamics.}
Figure~\ref{fig:experiment1_final}(b,c,h,i) illustrates the training dynamics of Avg@32 on AIME2024 and AIME2025. DGPO consistently outperforms all baselines in the mid-to-late training stages. Notably, algorithms with divergent gradient weights at the left boundary (CISPO, GPPO, and CE-GPPO) succumb to training collapse, and algorithms with larger policy gradient bias (GRPO and ASPO) exhibit suboptimal convergence.

Figure~\ref{fig:experiment1_final}(d,j) depicts the entropy dynamics. GRPO exhibits an entropy drop early in the training process, indicating premature exploitation that limits the achievable performance ceiling. Conversely, ASPO maintains excessively high entropy, reflecting over-exploration and insufficient exploitation, while CISPO, GPPO, and CE-GPPO show unstable entropy patterns leading to eventual collapse. DGPO demonstrates a moderate and controlled entropy reduction, signifying an optimal balance between exploration and exploitation.

\subsection{Hyperparameter Analysis}
\label{sec:hyper_analysis}

\begin{figure*}[htbp]
\centering
 \includegraphics[width=1.0\linewidth]{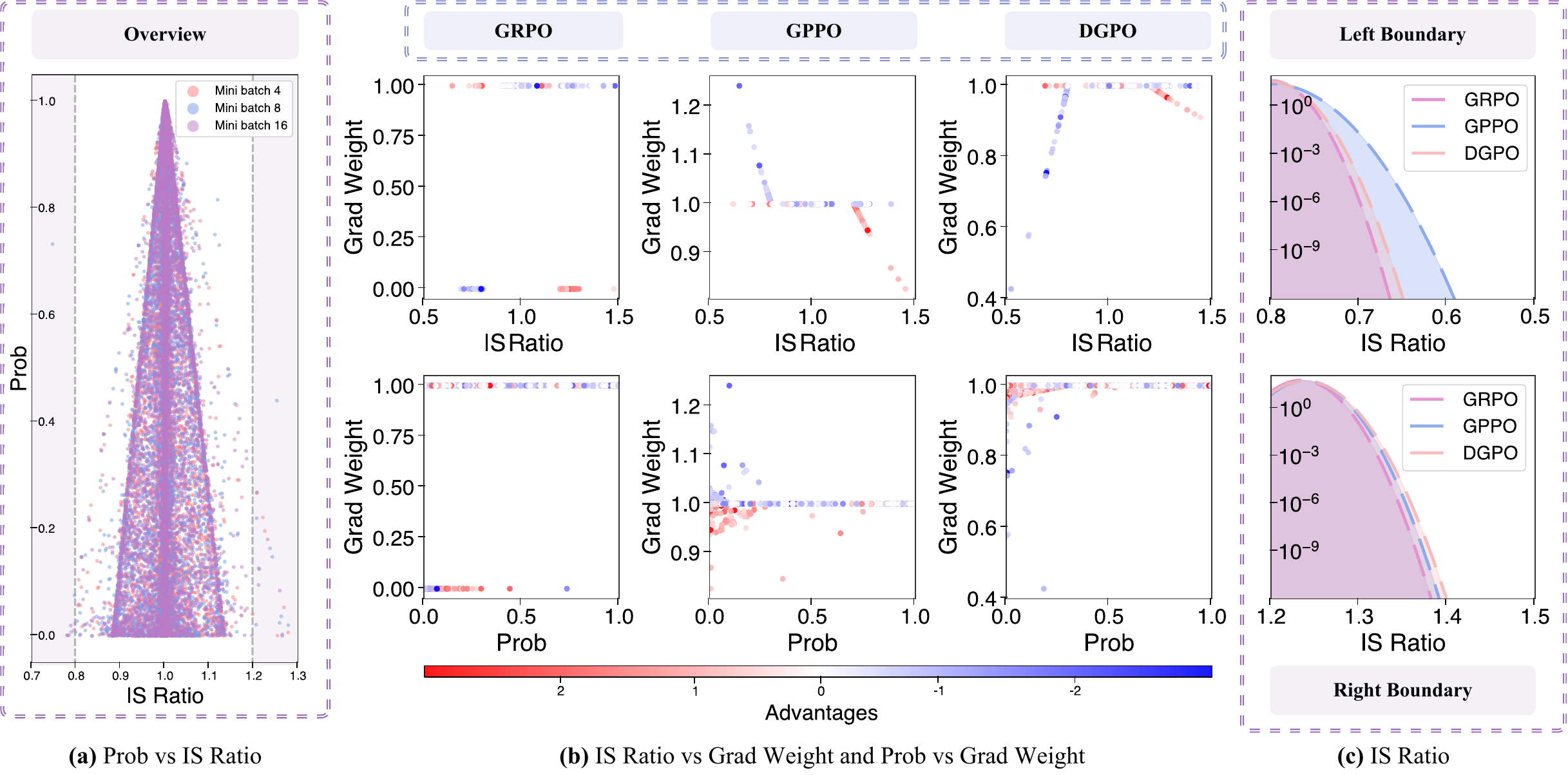}
\caption{Comparison of gradient weight distributions (Prob: Probability, Grad: Gradient). (a) Overall distribution. (b) Detailed scatter plots of grad weight vs. IS ratio (top row) and prob (bottom row) across three methods: GRPO (left), GPPO (middle), and DGPO (right). Points are colored by advantage. (c) Boundary distribution analysis.}
\label{fig:experiment2}
\vspace{-0.2em}
\end{figure*}

We investigate the impact of hyperparameters $n$ and $m$ by expanding from the baseline ($n=1$, $m=1$) to four configurations: $(1,1)$, $(1,2)$, $(2,1)$, and $(2,2)$. Figure \ref{fig:experiment1_final}(e,k) shows their performance on AIME25 across 1.5B and 7B models. 

\paragraph{Robustness and Patterns.}
All DGPO configurations outperform GRPO (with the exception of $n=2$ and $m=2$ on the 7B model), demonstrating our algorithm's robustness. However, the optimal configuration varies by scale: $(2,2)$ for 1.5B and $(1,2)$ for 7B. Analyzing Figure~\ref{fig:experiment1_final}(e,f,k,l), which reveals a consistent pattern: increasing $n$ or $m$ generally yields: (1) improved performance, (2) elevated overall entropy levels, but (3) reduced entropy stability. Notably, significant entropy volatility is observed only in the 7B model under the $(2,2)$ setting that negates the performance benefits.

\paragraph{Tuning Guideline.}
Based on these observations, we propose a heuristic for hyperparameter tuning: \textit{Enhance exploration by increasing $n$ and $m$ as long as entropy remains stable. Upon observing instability, revert to the preceding stable configuration.} Empirically, we recommend $n=1$ and $m=2$ as a robust and conservative baseline configuration.

\begin{table}[h]
\vspace{-2pt}
\centering
\small
\renewcommand{\arraystretch}{1.6} 
\setlength{\tabcolsep}{5.3pt}       
\caption{Scalability analysis of average Avg@32 (A) and Pass@32 (P) across 1.5B, 7B, and 14B models.}
\begin{tabular}{lccc}
\toprule
\multirow{2}{*}{\textbf{Method}} & \textbf{1.5B} & \textbf{7B} & \textbf{14B}\footnotemark \\ 
 & (A / P) & (A / P) & (A / P) \\
\midrule
GRPO & 48.4 / 70.1 & 58.9 / 77.2 & 53.6 / 67.4 \\
\rowcolor{ourshighlight}
\textbf{DGPO} & \best{52.7} / \best{73.0} & \best{62.0} / \best{77.9} & \best{56.7} / \best{70.4} \\
\textit{Improvement} & \improve{+4.3} / \improve{+2.9} & \improve{+3.1} / \improve{+0.7} & \improve{+3.1} / \improve{+3.0} \\
\bottomrule
\end{tabular}
\label{tab:scaling_results}
\vspace{-8pt}
\end{table}

\footnotetext{The performance fluctuation is attributed to the base model difference: 1.5B/7B models are derived from Qwen-2.5-Math, whereas the 14B model is derived from Qwen-2.5-Base.}

\paragraph{Scaling to Larger Models.}
The instability of $(2,2)$ on 7B (absent in 1.5B) suggests that larger models exhibit higher intrinsic entropy volatility, necessitating more conservative hyperparameters. To validate this pattern, we apply the optimal 7B configuration ($n=1, m=2$) to the 14B model. 

As summarized in Table~\ref{tab:scaling_results}, DGPO consistently outperforms GRPO across all model scales in both Avg@32 and Pass@32, which confirms that the benefits of our decoupled gradient mechanism effectively transfer to larger models. Furthermore, the training dynamics and entropy variations of the 14B model (shown in Appendix \ref{app:14b_dynamics}) exhibit stable convergence patterns similar to 7B under DGPO.

\subsection{Mechanistic Analysis via Visualization}
\paragraph{Mechanism of Stability (Left Boundary).}
\textit{Why does DGPO prevent collapse?}
Figure \ref{fig:experiment2}(a) visualizes the joint distribution of probability and IS ratios in the final mini-batch. Crucially, tokens at both boundaries are predominantly low-probability tokens. Figure \ref{fig:experiment2}(b) further details the relationship between probability, IS ratio, advantage, and relative gradient weight (normalized by $\pi_{\theta_\text{old}}$ probability) for GRPO (zero weight), GPPO (divergent weight), and DGPO (convergent weight). As illustrated, GRPO's zero gradients at the left boundary lead to the narrowest ratio distribution and insufficient exploration. Conversely, GPPO's divergent weights induce an excessively broad distribution, eventually precipitating training collapse due to instability (Figure \ref{fig:experiment1_final}(b,c,h,i))~\cite{yang2025not}. In contrast, DGPO maintains convergent relative weights, resulting in a ratio distribution only slightly wider than GRPO that successfully balances stability with effective exploration.

\paragraph{Mechanism of Improvement (Right Boundary).}
\textit{Why does DGPO perform better?}
Compared with GRPO (where right-boundary tokens have zero gradient and the narrowest ratio distribution), DGPO maintains gradients to foster exploration. 
Compared with GPPO (which uses a reciprocal standard weight equivalent to DGPO's $m=1$), DGPO with $m=2$ employs a reciprocal radical weight. This design induces the widest ratio distribution on the right boundary (Figure \ref{fig:experiment2}(c)), significantly enhancing performance. This finding aligns with prior research suggesting that increasing the contribution of positive samples improves performance~\cite{yang2025not, xi2025bapo}, and is consistent with our hyperparameter analysis in Section \ref{sec:hyper_analysis} where increasing $m$ boosts results.

\section{Conclusion}
In this paper, we revisit the fundamental optimization primitive in RLVR and establish probability—rather than log-probability—as the essential alignment target. We argue that the prevalent focus on log-probability gradients in prior ``soft clipping'' methods constitutes a misalignment with the true RL objective, leading to divergent gradient weights at boundaries and consequent training instability. To address this critical issue, we propose DGPO, which directly optimizes probability gradients via a decoupled decay mechanism based on IS ratios. This approach effectively resolves the inherent conflict between training stability and gradient preservation. Extensive experiments across the DeepSeek-R1-Distill-Qwen series models with multiple scales (1.5B, 7B, and 14B) demonstrate that DGPO consistently outperforms competitive baselines on various mathematical reasoning benchmarks, validating that aligning with the probability objective is crucial for unlocking the full potential of LLM reasoning capabilities through RL.

\section{Limitations}
We discuss two practical limitations of our work. (1) \textbf{Domain Specificity:} In line with most RLVR studies, our experiments are primarily conducted on mathematical reasoning datasets (e.g., AIME, MATH500) where verifiable rewards are readily available. While we believe the principles of probability gradients are generalizable, the efficacy of DGPO in domains with sparse or subjective rewards (e.g., creative writing) remains to be verified. We anticipate that the principles underlying DGPO generalize beyond the current setting: applying them to adjacent challenges—such as training data quality improvement~\cite{liu2023retrieval, liu2024unsupervised, liu2025stole}, parameter-efficient fine-tuning~\cite{dong2025aurora}, structured reasoning~\cite{dong2026neureasonerexplainablecontrollableunified, jiang2026foeforesterrorsmakes, zhang2026semanticawarelogicalreasoningsemiotic}, agentic exploration~\cite{zhang2026expseek}, and non-textual domains~\cite{HABIT, OFFSET, INTENT}—constitutes a promising direction for future work. (2) \textbf{Computational Constraints:} Due to limited computational resources, our scaling laws analysis is restricted to models up to 14B parameters. Although we observed consistent performance gains and predictable hyperparameter patterns from 1.5B to 14B, validating these findings on significantly larger foundation models (e.g., 70B+) would provide further insights into the method's scalability.

\section{Ethical Considerations}
We have carefully considered the ethical implications of our research and provide the following statements: (1) \textbf{Compliance and Transparency:} Throughout this study, we have strictly followed established ethical guidelines. Our findings are reported honestly, and we provide comprehensive theoretical derivations to ensure transparency. (2) \textbf{Data Usage:} The datasets employed in our experiments (e.g., DAPO-Math-17k) originate from publicly available sources. No private, sensitive, or confidential user information was used at any stage of our research. (3) \textbf{Reproducibility:} We offer detailed descriptions of the hyperparameter configurations, including the specific tuning guidelines for $n$ and $m$, to ensure the reproducibility of our results. (4) \textbf{Open Source:} In the interest of openness and to facilitate future research in the RLVR community, we have made our code available anonymously and will fully open-source it upon the acceptance of this paper.

\bibliography{references}

\appendix

\section{Derivation and Proof}

\subsection{Derivation of RL Estimator in SFT Form}
\label{app:rl_sft_form_derivation}

The token-level objective of RL is to maximize the arithmetic mean expectation of the estimated advantage, formulated as follows:
\begin{flalign}
\mathcal{J_\text{RL}}(\theta) = \mathbb{E}_{q \sim \mathcal{D},\ o \sim \pi_\theta(\cdot|q)} \left[ \sum_{t=1}^{|o|} A_t \right]
\end{flalign}
Assume that $\pi_e(\cdot|q)$ represents the ideal expert distribution, which satisfies two key properties: (1) its output follows a one-hot distribution; (2) every sampled token $o_t$ achieves the maximum advantage $A_t$. By applying Importance Sampling, we obtain:
\begin{align}
\mathcal{J_\text{RL}}(\theta) =& \mathbb{E}_{q \sim \mathcal{D},\ o \sim \pi_e(\cdot|q)} & \nonumber \\ 
&\left[ \sum_{t=1}^{|o|} \frac{\pi_\theta(o_t \mid q, o_{<t})}{\pi_e(o_t \mid q, o_{<t})} \cdot A_t \right]
\end{align}
Since $\pi_e(\cdot|q)$ is a one-hot distribution, the probability $\pi_e(o_t \mid q, o_{<t})$ is always 100\%, leading to:
\begin{align}
\mathcal{J_\text{RL}}(\theta) =& \mathbb{E}_{q \sim \mathcal{D},\ o \sim \pi_e(\cdot|q)} & \nonumber \\
&\left[ \sum_{t=1}^{|o|} \pi_\theta(o_t \mid q, o_{<t}) \cdot A_t \right]
\end{align}
Given that each action $o_t$ attains the highest possible $A_t$, under the binary advantage setting, $A_t$ is consistently 1, resulting in:
\begin{align}
\mathcal{J_\text{RL}}(\theta) = \mathbb{E}_{(q,o) \sim \mathcal{D}_e} \left[\sum_{t=1}^{|o|}\pi_\theta(o_t \mid q, o_{<t}) \right]
\end{align}
Consequently, the gradient is derived as:
\begin{align}
\nabla_\theta\mathcal{J_\text{RL}}(\theta) =& \mathbb{E}_{(q,o) \sim \mathcal{D}_e} & \nonumber \\
&\left[\sum_{t=1}^{|o|} \nabla_\theta \pi_\theta(o_t \mid q, o_{<t}) \right]
\end{align}

\subsection{Derivation of Continuity Constants}
\label{app:continuity_proof}

In this section, we derive the constants $C_{\text{left}}$ and $C_{\text{right}}$ used in the DGPO weighting function. The primary objective is to ensure that the gradient estimator is continuous with respect to the policy probability $\pi_\theta$ at the clipping boundaries.

From equation \eqref{equ:dgpo_loss}, the gradient equation of DGPO can be explicitly expressed as:
\begin{flalign}
&\nabla_\theta \mathcal{J}_{\text{DGPO}}(\theta) = \mathbb{E} _{q \sim \mathcal{D}, \{o_i\}_{i=1}^G \sim \pi_{\theta_{\text{old}}}(\cdot|q)} \frac{1}{\sum_{i=1}^G |o_i|} & \nonumber \\
&\sum_{i=1}^G \sum_{t=1}^{|o_i|} \mathcal{W}^\text{DGPO}_{i,t}(\theta) \hat{A}_{i} \nabla_\theta \pi_\theta(o_{i,t}|q,o_{i,<t}),
\end{flalign}
Using the identity $\nabla_\theta \pi_\theta = \pi_\theta \nabla_\theta \log \pi_\theta$, the effective coefficient applied to the standard score function $\nabla_\theta \log \pi_\theta$ is $\mathcal{F}_{i,t} = \mathcal{W}^\text{DGPO}_{i,t} \cdot \pi_\theta$. Since $\pi_\theta$ is continuous, ensuring the continuity of $\mathcal{W}^\text{DGPO}_{i,t}$ at the boundaries is sufficient to ensure the continuity of the entire gradient estimator.

Let $w_{i,t} = \frac{\pi_\theta}{\pi_{\theta_{\text{old}}}}$ denote the importance sampling ratio. The In-Boundary (M) weight is given by:
\begin{equation}
\mathcal{W}^{\text{M}}_{i,t} = \frac{1}{\pi_{\theta_{\text{old}}}}.
\end{equation}

\subsubsection{Left Boundary Derivation}
The transition between the Left Boundary (LN) and the In-Boundary (M) region occurs when the importance sampling ratio is $w_{i,t} = 1 - \varepsilon_{\text{low}}$. At this boundary, the current policy probability is:
\begin{equation}
\pi_\theta = (1 - \varepsilon_{\text{low}}) \pi_{\theta_{\text{old}}}.
\end{equation}
We then equate the weighting functions for the LN and M regions at this point:
\begin{align}
\mathcal{W}^{\text{LN}}_{i,t} \Big|_{w_{i,t}=1-\varepsilon_{\text{low}}} &= \mathcal{W}^{\text{M}}_{i,t} \Big|_{w_{i,t}=1-\varepsilon_{\text{low}}} & \nonumber \\
C_{\text{left}} \cdot \pi_\theta^n &= \frac{1}{\pi_{\theta_{\text{old}}}} & \nonumber \\
C_{\text{left}} \cdot \left[ (1 - \varepsilon_{\text{low}}) \pi_{\theta_{\text{old}}} \right]^n &= \frac{1}{\pi_{\theta_{\text{old}}}}.
\end{align}
Solving for $C_{\text{left}}$:
\begin{align}
C_{\text{left}} &= \frac{1}{\pi_{\theta_{\text{old}}} \cdot (1 - \varepsilon_{\text{low}})^n \pi_{\theta_{\text{old}}}^n} & \nonumber \\
&=\frac{1}{(1 - \varepsilon_{\text{low}})^n \pi_{\theta_{\text{old}}}^{n+1}}.
\end{align}
Thus, $C_{\text{left}} = (1 - \varepsilon_{\text{low}})^{-n} \pi_{\theta_{\text{old}}}^{-(n+1)}$.

\begin{table*}[htbp]
    \centering
    \setlength{\tabcolsep}{10pt}
    \caption{Definition of bias relative to Policy Gradient}
    \begin{tabular}{l c}
        \toprule
        \textbf{Bias Type} & \textbf{Mathematical Definition}\\
        \midrule
        In-Boundary Bias &
        $\text{Bias}_\text{Algo}^\text{M}=\left\| \nabla_\theta\mathcal{J}_{\text{Algo}}^\text{M}(\theta) - \nabla_\theta\mathcal{J}_{\text{PG}}^\text{M}(\theta) \right\|$ \\
        \addlinespace[2pt]
        Left-Boundary Bias &
        $\text{Bias}_\text{Algo}^\text{LN}=\left\| \nabla_\theta\mathcal{J}_{\text{Algo}}^\text{LN}(\theta) - \nabla_\theta\mathcal{J}_{\text{PG}}^\text{LN}(\theta) \right\|$ \\
        \addlinespace[2pt]
        Right-Boundary Bias &
        $\text{Bias}_\text{Algo}^\text{HP}=\left\| \nabla_\theta\mathcal{J}_{\text{Algo}}^\text{HP}(\theta) - \nabla_\theta\mathcal{J}_{\text{PG}}^\text{HP}(\theta) \right\|$ \\
        \addlinespace[2pt]
        Reverse Left-Boundary Bias &
        $\text{Bias}_\text{Algo}^\text{LP}=\left\| \nabla_\theta\mathcal{J}_{\text{Algo}}^\text{LP}(\theta) - \nabla_\theta\mathcal{J}_{\text{PG}}^\text{LP}(\theta) \right\|$ \\
        \addlinespace[2pt]
        Reverse Right-Boundary Bias &
        $\text{Bias}_\text{Algo}^\text{HN}=\left\| \nabla_\theta\mathcal{J}_{\text{Algo}}^\text{HN}(\theta) - \nabla_\theta\mathcal{J}_{\text{PG}}^\text{HN}(\theta) \right\|$ \\
        \bottomrule
    \end{tabular}
    \label{tab:bias_definition}
\end{table*}

\subsubsection{Right Boundary Derivation}
Similarly, the transition between the Right Boundary (HP) and the In-Boundary (M) region occurs when $w_{i,t} = 1 + \varepsilon_{\text{high}}$. At this boundary:
\begin{equation}
\pi_\theta = (1 + \varepsilon_{\text{high}}) \pi_{\theta_{\text{old}}}.
\end{equation}
We similarly equate the weighting functions for the HP and M regions at this point:
\begin{align}
\mathcal{W}^{\text{HP}}_{i,t} \Big|_{w_{i,t}=1+\varepsilon_{\text{high}}} &= \mathcal{W}^{\text{M}}_{i,t} \Big|_{w_{i,t}=1+\varepsilon_{\text{high}}} & \nonumber \\
C_{\text{right}} \cdot \pi_\theta^{-\frac{1}{m}} &= \frac{1}{\pi_{\theta_{\text{old}}}} & \nonumber \\
C_{\text{right}} \cdot \left[ (1 + \varepsilon_{\text{high}}) \pi_{\theta_{\text{old}}} \right]^{-\frac{1}{m}} &= \frac{1}{\pi_{\theta_{\text{old}}}}.
\end{align}
Solving for $C_{\text{right}}$:
\begin{align}
C_{\text{right}} &= \frac{(1 + \varepsilon_{\text{high}})^{\frac{1}{m}} \pi_{\theta_{\text{old}}}^{\frac{1}{m}}}{\pi_{\theta_{\text{old}}}} & \nonumber \\
&= (1 + \varepsilon_{\text{high}})^{\frac{1}{m}} \pi_{\theta_{\text{old}}}^{\frac{1}{m}-1}.
\end{align}
This concludes the derivation of the constants.

\subsection{Proof of Policy Gradient Bias}
\label{app:bias_proof}

\subsubsection{Standard Policy Gradient Estimator}
The gradient estimator for the standard policy gradient method is formally given by:
\begin{align}
\nabla_\theta \mathcal{J}_\text{PG}(\theta) =& \mathbb{E}_{q \sim \mathcal{D},\ o \sim \pi_\theta(\cdot|q)} & \nonumber \\
&\left[ A_t \nabla_\theta \log\pi_\theta(o_t | q, o_{<t}) \right]
\end{align}
By introducing Importance Sampling with the old policy, we obtain the following form:
\begin{align}
\label{equ:standard_pg}
&\nabla_\theta \mathcal{J}_\text{PG}(\theta) = \mathbb{E}_{q \sim \mathcal{D}, o \sim \pi_{\theta_\text{old}}(\cdot|q)} & \nonumber \\
& \quad \left[ \frac{\pi_\theta(o_t | q, o_{<t})}{\pi_{\theta_\text{old}}(o_t | q, o_{<t})} A_t \nabla_\theta \log\pi_\theta(o_t | q, o_{<t}) \right]
\end{align}
We adopt Equation \eqref{equ:standard_pg} as the standard policy gradient estimator in the subsequent proofs.

\subsubsection{Decomposed Gradient Estimation}
To distinguish gradient estimators under different conditions, we define the following binary variables based on the importance sampling ratio $r_t(\theta) = \frac{\pi_\theta(o_t | q, o_{<t})}{\pi_{\theta_\text{old}}(o_t | q, o_{<t})}$ and the advantage $A_t$. Note that while $r_t$ is a function of $\theta$, for the purpose of defining the estimator's functional form, these regions are treated as piecewise conditions:
\begin{flalign}
\begin{cases}
v_t^{\text{LN}} = \mathbb{I}\left[ r_t(\theta) < 1 - \varepsilon_{\text{low}} \land A_t < 0 \right] \\
v_t^{\text{HP}} = \mathbb{I}\left[ r_t(\theta) > 1 + \varepsilon_{\text{high}} \land A_t > 0 \right] \\
v_t^{\text{LP}} = \mathbb{I}\left[ r_t(\theta) < 1 - \varepsilon_{\text{low}} \land A_t > 0 \right] \\
v_t^{\text{HN}} = \mathbb{I}\left[ r_t(\theta) > 1 + \varepsilon_{\text{high}} \land A_t  < 0 \right] \\
v_t^{\text{M}} = 1 - \left( v_t^{\text{LN}} + v_t^{\text{HP}} + v_t^{\text{LP}} + v_t^{\text{HN}} \right)
\end{cases}.
\end{flalign}
The policy gradient estimator is formally decomposed and expressed as a sum of five distinct terms:
\begin{flalign}
\nabla_\theta \mathcal{J}_{\text{PG}}(\theta) =  \sum_{X \in \mathcal{X}} \nabla_\theta\mathcal{J}_{\text{PG}}^X(\theta),
\end{flalign}
where $ \mathcal{X} \in \{\text{M, LN, LP, HP, HN}\}$, and the general mathematical form for each term is defined as:
\begin{align}
&\nabla_\theta\mathcal{J}_{\text{PG}}^X(\theta) = \mathbb{E}_{q \sim \mathcal{D},\, o \sim \pi_{\theta_{\text{old}}}(\cdot|q)} & \nonumber \\
& \qquad \left[ v_t^{X} \cdot r_t  A_t \nabla_\theta \log \pi_\theta(o_t | q, o_{<t}) \right].
\end{align}
Similarly, gradient estimators for GRPO, CISPO, GPPO, CE-GPPO, ASPO, and DGPO can be consistently decomposed into five-term sums. The bias is defined as the magnitude ($L_2$ norm) of the difference vector between the algorithm's gradient estimator and the standard policy gradient under identical conditions, as shown in Table \ref{tab:bias_definition}.

\subsubsection{In-Boundary Bias}
The gradient estimates for each algorithm under in-boundary conditions are analyzed below. While most algorithms (GRPO, CISPO, GPPO, CE, DGPO) maintain an unbiased estimator locally ($v_t^M=1$), ASPO introduces a non-linear regularization term specifically when the advantage is positive ($A_t > 0$).
The magnitude of the bias is then computed as shown in Table \ref{tab:M_bias}.

\begin{table*}[htbp]
    \centering
    \small
    \caption{In-Boundary bias among various policy optimization algorithms}
    \label{tab:M_bias}
    \begin{tabular}{l l}
        \toprule
        \textbf{Algorithm} & \textbf{In-Boundary Bias Magnitude}\\
        \midrule
        GRPO, CISPO, & \multirow{3}{*}{$\displaystyle \text{Bias}_\text{X}^\text{M} = \left\| \nabla_\theta\mathcal{J}_{\text{X}}^\text{M}(\theta) - \nabla_\theta\mathcal{J}_{\text{PG}}^\text{M}(\theta) \right\| = 0$} \\
        GPPO, CE-GPPO & \\
        DGPO & \\
        \addlinespace[6pt]
        \multirow{2}{*}{ASPO} &
        $\begin{aligned}
            \text{Bias}_\text{ASPO}^\text{M} &= \left\| \mathbb{E}_{q \sim \mathcal{D},\, o \sim \pi_{\theta_{\text{old}}}(\cdot|q)} \left[ v_t^{\text{M}} \cdot \left( \frac{1}{r_t} - r_t \right) \cdot A_t \nabla_\theta \log \pi_\theta(o_t | q, o_{<t}) \right] \right\| \\
            &\approx \left\| \mathbb{E}_{q \sim \mathcal{D},\, o \sim \pi_{\theta_{\text{old}}}(\cdot|q)} \left[ v_t^{\text{M}} \cdot \underbrace{-2(r_t-1)}_{\lambda_t} \cdot A_t \nabla_\theta \log \pi_\theta(o_t | q, o_{<t}) \right] \right\| \neq 0^*
        \end{aligned}$ \\
        \bottomrule
    \end{tabular}
\end{table*}

\noindent The relationship of in-boundary bias magnitudes is determined to be: 
\begin{equation*}
0 = \textcolor{biascolor}{\text{Bias}_\text{DGPO}^\text{M}} = \text{Bias}_\text{Others}^\text{M} < \text{Bias}_\text{ASPO}^\text{M}.
\end{equation*}

\subsubsection{Left-Boundary Bias}
The gradient estimates for each algorithm under specific left-boundary conditions ($r_t < 1-\varepsilon_\text{low}, A_t < 0$) are presented in Eq. \eqref{eq:LB_gradients}. Let $r_0=1-\varepsilon_\text{low}$. The magnitude of the bias is computed as shown in Table \ref{tab:LN_bias}.

Since $0 < r_t < r_0< 1$, it follows universally that $\text{Bias}_\text{DGPO}^\text{LN} < \text{Bias}_\text{GRPO}^\text{LN}$ and $\text{Bias}_\text{DGPO}^\text{LN} < \text{Bias}_\text{ASPO}^\text{LN}$. However, ranking other algorithms requires a quantitative analysis of the integrals.

\paragraph{Decoupling Assumption and Quantitative Analysis.}
To rigorously compare the bias magnitudes, we apply a \textbf{decoupling assumption}: we assume the gradient norm is locally independent of the importance sampling ratio $r_t$ within the small boundary region. Let $\delta = \mathbb{E}_{q \sim \mathcal{D}}[\lVert \nabla_\theta \log \pi_\theta(o_t | q, o_{<t}) \rVert]$ denote the average gradient magnitude. 

The explicit expressions analytically  are derived as follows (integrals are over $r_t$):

\begin{align}
\label{equ:left_bias1}
\text{Bias}_\text{GRPO}^\text{LN} &= \text{Bias}_\text{ASPO}^\text{LN} \notag \\
&= \int_0^{r_0} \underbrace{|0 - r_t|}_{\text{Coeff. Diff}} \cdot \delta \cdot k r_t^\gamma \, \mathrm{d}r_t \notag \\
&= \frac{k \delta r_0^{\gamma+2}}{\gamma+2}
\end{align}

\begin{align}
\label{equ:left_bias2}
\text{Bias}_\text{CISPO}^\text{LN} &= \text{Bias}_\text{GPPO}^\text{LN} \notag \\
&=\int_0^{r_0} \underbrace{|r_0 - r_t|}_{\text{Coeff. Diff}} \cdot \delta \cdot k r_t^\gamma \, \mathrm{d}r_t \notag \\
&= k \delta r_0^{\gamma+2} \left( \frac{1}{\gamma+1} - \frac{1}{\gamma+2} \right)
\end{align}

\begin{align}
\label{equ:left_bias3}
\text{Bias}_\text{CE}^\text{LN} &= \int_0^{r_0} \underbrace{|\beta_1 r_0 - r_t|}_{\text{Coeff. Diff}} \cdot \delta \cdot k r_t^\gamma \, \mathrm{d}r_t \notag \\
&= k \delta r_0^{\gamma+2} \left| \frac{\beta_1}{\gamma+1} - \frac{1}{\gamma+2} \right|
\end{align}

\begin{align}
\label{equ:left_bias4}
\text{Bias}_\text{DGPO}^\text{LN} &=\int_0^{r_0} \underbrace{\left| \frac{r_t^{n+1}}{r_0^n} - r_t \right|}_{\text{Coeff. Diff}} \cdot \delta \cdot k r_t^\gamma \, \mathrm{d}r_t \notag \\
&= k \delta r_0^{\gamma+2} \left( \frac{1}{\gamma+2} - \frac{1}{n+\gamma+2} \right)
\end{align}
where we utilize the property $r_t > r_t^{n+1}/r_0^n$ for $r_t < r_0$.
In the extreme limiting case ($\gamma \to +\infty$), the resulting bias ratios are:
\begin{align}
\label{equ:lim1}
\lim_{\gamma \to \infty}\frac{\text{Bias}_{\text{DGPO}}^{\text{LN}}}{\text{Bias}_{\text{CISPO}}^{\text{LN}}} &= \lim_{\gamma \to \infty}\frac{\text{Bias}_{\text{DGPO}}^{\text{LN}}}{\text{Bias}_{\text{GPPO}}^{\text{LN}}} \notag \\
&= \lim_{\gamma \to \infty} \frac{n(\gamma+1)}{\gamma+2+n} = n
\end{align}
\begin{align}
\label{equ:lim2}
\lim_{\gamma \to \infty}\frac{\text{Bias}_{\text{DGPO}}^{\text{LN}}}{\text{Bias}_{\text{CE}}^{\text{LN}}} &= \lim_{\gamma \to \infty} \frac{n(\gamma+1)(\gamma+2+n)^{-1}}{\left| (1-\beta_1)\gamma+1-2\beta_1 \right|} \notag \\
&= 0
\end{align}
\begin{align}
\label{equ:lim3}
\lim_{\gamma \to \infty}\frac{\text{Bias}_{\text{CISPO}}^{\text{LN}}}{\text{Bias}_{\text{CE}}^{\text{LN}}} &= \lim_{\gamma \to \infty}\frac{\text{Bias}_{\text{GPPO}}^{\text{LN}}}{\text{Bias}_{\text{CE}}^{\text{LN}}} \notag \\
&= \lim_{\gamma \to \infty} \frac{1}{\left| (1-\beta_1)\gamma+1-2\beta_1 \right|} & \nonumber \\
&= 0
\end{align}
\begin{align}
\label{equ:lim4}
\lim_{\gamma \to \infty}\frac{\text{Bias}_{\text{GRPO}}^{\text{LN}}}{\text{Bias}_{\text{CE}}^{\text{LN}}} &= \lim_{\gamma \to \infty}\frac{\text{Bias}_{\text{ASPO}}^{\text{LN}}}{\text{Bias}_{\text{CE}}^{\text{LN}}} \notag \\
&= \lim_{\gamma \to \infty} \frac{\gamma+1}{\left| (1-\beta_1)\gamma+1-2\beta_1 \right|} \notag \\
&= \frac{1}{|1-\beta_1|}
\end{align}
Assuming $n=1$ (linear decay), we obtain the strict ranking:
$0 < \textcolor{biascolor}{\text{Bias}_\text{DGPO}^\text{LN}} < \text{Bias}_\text{CISPO}^\text{LN} = \text{Bias}_\text{GPPO}^\text{LN} < \text{Bias}_\text{CE}^\text{LN} < \text{Bias}_\text{GRPO}^\text{LN} = \text{Bias}_\text{ASPO}^\text{LN}$.

\subsubsection{Right-Boundary Bias}
The gradient estimates for the right-boundary case across algorithms are shown in Eq. \eqref{eq:RB_gradients}. The bias magnitude is computed as shown in Table \ref{tab:HP_bias}.

\begin{table*}[!h]
    \centering
    \small
    \caption{Left-Boundary bias among various policy optimization algorithms}
    \label{tab:LN_bias}
    \begin{tabular}{l l}
        \toprule
        \textbf{Algorithm} & \textbf{Left-Boundary Bias Magnitude}\\
        \midrule
        GRPO &
        $\begin{aligned}[t]
            \text{Bias}_\text{GRPO}^\text{LN} &= \left\| \nabla_\theta\mathcal{J}_{\text{GRPO}}^\text{LN}(\theta) - \nabla_\theta\mathcal{J}_{\text{PG}}^\text{LN}(\theta) \right\| \\
            &= \left\| \mathbb{E}_{q \sim \mathcal{D},\, o \sim \pi_{\theta_{\text{old}}}} \left[ v_t^{\text{LN}} \cdot r_t \cdot A_t \nabla_\theta \log \pi_\theta(o_t | q, o_{<t}) \right] \right\|
        \end{aligned}$ \\
        \addlinespace[4pt]
        CISPO &
        $\begin{aligned}[t]
            \text{Bias}_\text{CISPO}^\text{LN} &= \left\| \nabla_\theta\mathcal{J}_{\text{CISPO}}^\text{LN}(\theta) - \nabla_\theta\mathcal{J}_{\text{PG}}^\text{LN}(\theta) \right\| \\
            &= \left\| \mathbb{E}_{q \sim \mathcal{D},\, o \sim \pi_{\theta_{\text{old}}}} \left[ v_t^{\text{LN}} \cdot (r_0-r_t) \cdot A_t \nabla_\theta \log \pi_\theta(o_t | q, o_{<t}) \right] \right\|
        \end{aligned}$ \\
        \addlinespace[4pt]
        GPPO &
        $\begin{aligned}[t]
            \text{Bias}_\text{GPPO}^\text{LN} &= \left\| \nabla_\theta\mathcal{J}_{\text{GPPO}}^\text{LN}(\theta) - \nabla_\theta\mathcal{J}_{\text{PG}}^\text{LN}(\theta) \right\| \\
            &= \left\| \mathbb{E}_{q \sim \mathcal{D},\, o \sim \pi_{\theta_{\text{old}}}} \left[ v_t^{\text{LN}} \cdot (r_0-r_t) \cdot A_t \nabla_\theta \log \pi_\theta(o_t | q, o_{<t}) \right] \right\|
        \end{aligned}$ \\
        \addlinespace[4pt]
        CE-GPPO &
        $\begin{aligned}[t]
            \text{Bias}_\text{CE}^\text{LN} &= \left\| \nabla_\theta\mathcal{J}_{\text{CE}}^\text{LN}(\theta) - \nabla_\theta\mathcal{J}_{\text{PG}}^\text{LN}(\theta) \right\| \\
            &= \left\| \mathbb{E}_{q \sim \mathcal{D},\, o \sim \pi_{\theta_{\text{old}}}} \left\{ v_t^{\text{LN}} \cdot [\beta_1 r_0-r_t] \cdot A_t \nabla_\theta \log \pi_\theta(o_t | q, o_{<t}) \right\} \right\|
        \end{aligned}$ \\
        \addlinespace[4pt]
        \multirow{1}{*}ASPO &
        $\begin{aligned}[t]
            \text{Bias}_\text{ASPO}^\text{LN}& = \left\| \nabla_\theta\mathcal{J}_{\text{ASPO}}^\text{LN}(\theta) - \nabla_\theta\mathcal{J}_{\text{PG}}^\text{LN}(\theta) \right\| \\
            &= \left\| \mathbb{E}_{q \sim \mathcal{D},\, o \sim \pi_{\theta_{\text{old}}}} \left[ v_t^{\text{LN}} \cdot r_t \cdot A_t \nabla_\theta \log \pi_\theta(o_t | q, o_{<t}) \right] \right\|
        \end{aligned}$ \\
        \addlinespace[4pt]
        DGPO &
        $\begin{aligned}[t]
            \text{Bias}_\text{DGPO}^\text{LN} &= \left\| \nabla_\theta\mathcal{J}_{\text{DGPO}}^\text{LN}(\theta) - \nabla_\theta\mathcal{J}_{\text{PG}}^\text{LN}(\theta) \right\| \\
            &= \left\| \mathbb{E}_{q \sim \mathcal{D},\, o \sim \pi_{\theta_{\text{old}}}} \left[ v_t^{\text{LN}} \cdot (r_t - \frac{r_t^{n+1}}{r_0^n}) \cdot A_t \nabla_\theta \log \pi_\theta(o_t | q, o_{<t}) \right] \right\|
        \end{aligned}$ \\
        \bottomrule
    \end{tabular}
\end{table*}

\begin{figure*}[t]
\begin{equation} \label{eq:LB_gradients}
\begin{cases}
\nabla_\theta\mathcal{J}_{\text{PG}}^\text{LN}(\theta) = \mathbb{E}_{q \sim \mathcal{D},\, o \sim \pi_{\theta_{\text{old}}}} \left[ v_t^{\text{LN}} \cdot r_t \cdot A_t \nabla_\theta \log \pi_\theta(o_t | q, o_{<t})\right] \\
\nabla_\theta\mathcal{J}_{\text{GRPO}}^\text{LN}(\theta) = \mathbb{E}_{q \sim \mathcal{D},\, o \sim \pi_{\theta_{\text{old}}}} \left[ v_t^{\text{LN}}  \cdot 0 \cdot  A_t \nabla_\theta \log \pi_\theta(o_t | q, o_{<t}) \right] \\
\nabla_\theta\mathcal{J}_{\text{CISPO}}^\text{LN}(\theta) = \mathbb{E}_{q \sim \mathcal{D},\, o \sim \pi_{\theta_{\text{old}}}} \left[ v_t^{\text{LN}} \cdot (1-\varepsilon_\text{low}) \cdot A_t \nabla_\theta \log \pi_\theta(o_t | q, o_{<t}) \right] \\ 
\nabla_\theta\mathcal{J}_{\text{GPPO}}^\text{LN}(\theta) = \mathbb{E}_{q \sim \mathcal{D},\, o \sim \pi_{\theta_{\text{old}}}} \left[ v_t^{\text{LN}} \cdot (1-\varepsilon_\text{low}) \cdot A_t \nabla_\theta \log \pi_\theta(o_t | q, o_{<t}) \right] \\
\nabla_\theta\mathcal{J}_{\text{CE}}^\text{LN} (\theta) = \mathbb{E}_{q \sim \mathcal{D},\, o \sim \pi_{\theta_{\text{old}}}} \left[ v_t^{\text{LN}} \cdot \beta_1 (1-\varepsilon_\text{low}) \cdot A_t \nabla_\theta \log \pi_\theta(o_t | q, o_{<t}) \right] \\
\nabla_\theta\mathcal{J}_{\text{ASPO}}^\text{LN} (\theta) = \mathbb{E}_{q \sim \mathcal{D},\, o \sim \pi_{\theta_{\text{old}}}} \left[  v_t^{\text{LN}} \cdot 0 \cdot  A_t \nabla_\theta \log \pi_\theta(o_t | q, o_{<t}) \right] \\
\nabla_\theta\mathcal{J}_{\text{DGPO}}^\text{LN} (\theta) = \mathbb{E}_{q \sim \mathcal{D},\, o \sim \pi_{\theta_{\text{old}}}} \left[  v_t^{\text{LN}} \cdot \frac{r_t^{n+1}}{(1-\varepsilon_\text{low})^n} \cdot A_t \nabla_\theta \log \pi_\theta(o_t | q, o_{<t}) \right]
\end{cases}
\end{equation}
\vspace{-8pt}
\end{figure*}

Given the right-boundary condition $1 < 1+\varepsilon_\text{high} < r_t$, and adopting the CE-GPPO configuration $\beta_2=1$, we derive the ranking:
\begin{multline*}
0 < \textcolor{biascolor}{\text{Bias}_\text{DGPO}^\text{HP}} \leq \text{Bias}_\text{CISPO}^\text{HP} = \text{Bias}_\text{GPPO}^\text{HP} \\
= \text{Bias}_\text{CE}^\text{HP} < \text{Bias}_\text{GRPO}^\text{HP} = \text{Bias}_\text{ASPO}^\text{HP}.
\end{multline*}

\subsubsection{Reverse Left-Boundary Bias}
Based on the definition of ASPO, if the reverse left-boundary is treated as a clipping region $1-\varepsilon_\text{low}$, the gradient estimates are listed in Eq. \eqref{eq:RLB_gradients}. The bias magnitude is shown in Table \ref{tab:LP_bias}.

The reverse left-boundary bias magnitude relationship is directly derived as: 
\begin{multline*}
0 = \textcolor{biascolor}{\text{Bias}_\text{DGPO}^\text{LP}} = \text{Bias}_\text{GRPO}^\text{LP} = \text{Bias}_\text{GPPO}^\text{LP} \\
= \text{Bias}_\text{CE}^\text{LP} < \text{Bias}_\text{CISPO}^\text{LP} = \text{Bias}_\text{ASPO}^\text{LP}.
\end{multline*}

\begin{table*}[!h]
    \centering
    \small
    \caption{Right-Boundary bias among various policy optimization algorithms}
    \label{tab:HP_bias}
    \begin{tabular}{l l}
        \toprule
        \textbf{Algorithm} & \textbf{Right-Boundary Bias Magnitude}\\
        \midrule
        GRPO &
        $\begin{aligned}[t]
            \text{Bias}_\text{GRPO}^\text{HP} &= \left\| \nabla_\theta\mathcal{J}_{\text{GRPO}}^\text{HP}(\theta) - \nabla_\theta\mathcal{J}_{\text{PG}}^\text{HP}(\theta) \right\| \\
            &= \left\| \mathbb{E}_{q \sim \mathcal{D},\, o \sim \pi_{\theta_{\text{old}}}} \left[ v_t^{\text{HP}} \cdot r_t \cdot A_t \nabla_\theta \log \pi_\theta(o_t | q, o_{<t}) \right] \right\|
        \end{aligned}$ \\
        \addlinespace[4pt]
        CISPO &
        $\begin{aligned}[t]
            \text{Bias}_\text{CISPO}^\text{HP} &= \left\| \nabla_\theta\mathcal{J}_{\text{CISPO}}^\text{HP}(\theta) - \nabla_\theta\mathcal{J}_{\text{PG}}^\text{HP}(\theta) \right\| \\
            &= \left\| \mathbb{E}_{q \sim \mathcal{D},\, o \sim \pi_{\theta_{\text{old}}}} \left\{ v_t^{\text{HP}} \cdot [r_t - (1+\varepsilon_\text{high})] \cdot A_t \nabla_\theta \log \pi_\theta(o_t | q, o_{<t}) \right\} \right\|
        \end{aligned}$ \\
        \addlinespace[4pt]
        GPPO &
        $\begin{aligned}[t]
            \text{Bias}_\text{GPPO}^\text{HP} &= \left\| \nabla_\theta\mathcal{J}_{\text{GPPO}}^\text{HP}(\theta) - \nabla_\theta\mathcal{J}_{\text{PG}}^\text{HP}(\theta) \right\| \\
            &= \left\| \mathbb{E}_{q \sim \mathcal{D},\, o \sim \pi_{\theta_{\text{old}}}} \left\{ v_t^{\text{HP}} \cdot  [r_t - (1+\varepsilon_\text{high})] \cdot A_t \nabla_\theta \log \pi_\theta(o_t | q, o_{<t}) \right\} \right\|
        \end{aligned}$ \\
        \addlinespace[4pt]
        CE-GPPO &
        $\begin{aligned}[t]
            \text{Bias}_\text{CE}^\text{HP} &= \left\| \nabla_\theta\mathcal{J}_{\text{CE}}^\text{HP}(\theta) - \nabla_\theta\mathcal{J}_{\text{PG}}^\text{HP}(\theta) \right\| \\
            &= \left\| \mathbb{E}_{q \sim \mathcal{D},\, o \sim \pi_{\theta_{\text{old}}}} \left\{ v_t^{\text{HP}} \cdot  [r_t - \beta_2 (1+\varepsilon_\text{high})] \cdot A_t \nabla_\theta \log \pi_\theta(o_t | q, o_{<t}) \right\} \right\|
        \end{aligned}$ \\
        \addlinespace[4pt]
        \multirow{1}{*}ASPO &
        $\begin{aligned}[t]
            \text{Bias}_\text{ASPO}^\text{HP} &= \left\| \nabla_\theta\mathcal{J}_{\text{ASPO}}^\text{HP}(\theta) - \nabla_\theta\mathcal{J}_{\text{PG}}^\text{HP}(\theta) \right\| \\
            &= \left\| \mathbb{E}_{q \sim \mathcal{D},\, o \sim \pi_{\theta_{\text{old}}}} \left[ v_t^{\text{HP}} \cdot r_t \cdot A_t \nabla_\theta \log \pi_\theta(o_t | q, o_{<t}) \right] \right\|
        \end{aligned}$ \\
        \addlinespace[4pt]
        DGPO &
        $\begin{aligned}[t]
            \text{Bias}_\text{DGPO}^\text{HP} &= \left\| \nabla_\theta\mathcal{J}_{\text{DGPO}}^\text{HP}(\theta) - \nabla_\theta\mathcal{J}_{\text{PG}}^\text{HP}(\theta) \right\| \\
            &= \left\| \mathbb{E}_{q \sim \mathcal{D},\, o \sim \pi_{\theta_{\text{old}}}} \left[ v_t^{\text{HP}} \cdot [r_t - (1+\varepsilon_\text{high})^\frac{1}{m} r_t^{1-\frac{1}{m}} ] \cdot A_t \nabla_\theta \log \pi_\theta(o_t | q, o_{<t}) \right] \right\|
        \end{aligned}$ \\
        \bottomrule
    \end{tabular}
\end{table*}

\begin{figure*}[t]
\begin{equation} \label{eq:RB_gradients}
\begin{cases}
\nabla_\theta\mathcal{J}_{\text{PG}}^\text{HP}(\theta) = \mathbb{E}_{q \sim \mathcal{D},\, o \sim \pi_{\theta_{\text{old}}}} \left[ v_t^{\text{HP}} \cdot r_t \cdot A_t \nabla_\theta \log \pi_\theta(o_t | q, o_{<t})\right] \\
\nabla_\theta\mathcal{J}_{\text{GRPO}}^\text{HP}(\theta) = \mathbb{E}_{q \sim \mathcal{D},\, o \sim \pi_{\theta_{\text{old}}}} \left[ v_t^{\text{HP}}  \cdot 0 \cdot  A_t \nabla_\theta \log \pi_\theta(o_t | q, o_{<t}) \right] \\
\nabla_\theta\mathcal{J}_{\text{CISPO}}^\text{HP}(\theta) = \mathbb{E}_{q \sim \mathcal{D},\, o \sim \pi_{\theta_{\text{old}}}} \left[ v_t^{\text{HP}} \cdot (1+\varepsilon_\text{high}) \cdot A_t \nabla_\theta \log \pi_\theta(o_t | q, o_{<t}) \right] \\ 
\nabla_\theta\mathcal{J}_{\text{GPPO}}^\text{HP}(\theta) = \mathbb{E}_{q \sim \mathcal{D},\, o \sim \pi_{\theta_{\text{old}}}} \left[ v_t^{\text{HP}} \cdot (1+\varepsilon_\text{high}) \cdot A_t \nabla_\theta \log \pi_\theta(o_t | q, o_{<t}) \right] \\
\nabla_\theta\mathcal{J}_{\text{CE}}^\text{HP} (\theta) = \mathbb{E}_{q \sim \mathcal{D},\, o \sim \pi_{\theta_{\text{old}}}} \left[ v_t^{\text{HP}} \cdot \beta_2 (1+\varepsilon_\text{high}) \cdot A_t \nabla_\theta \log \pi_\theta(o_t | q, o_{<t}) \right] \\
\nabla_\theta\mathcal{J}_{\text{ASPO}}^\text{HP} (\theta) = \mathbb{E}_{q \sim \mathcal{D},\, o \sim \pi_{\theta_{\text{old}}}} \left[  v_t^{\text{HP}} \cdot 0 \cdot  A_t \nabla_\theta \log \pi_\theta(o_t | q, o_{<t}) \right] \\
\nabla_\theta\mathcal{J}_{\text{DGPO}}^\text{HP} (\theta) = \mathbb{E}_{q \sim \mathcal{D},\, o \sim \pi_{\theta_{\text{old}}}} \left[  v_t^{\text{HP}} \cdot (1+\varepsilon_\text{high})^{\frac{1}{m}} r_t^{1-\frac{1}{m}} \cdot A_t \nabla_\theta \log \pi_\theta(o_t | q, o_{<t}) \right]
\end{cases}
\end{equation}
\end{figure*}

\subsubsection{Reverse Right-Boundary Bias}
Assuming the reverse right-boundary of ASPO is $1+\varepsilon_\text{high}$, the gradient estimates are listed in Eq. \eqref{eq:RRB_gradients}. The bias magnitude is shown in Table \ref{tab:HN_bias}.

The reverse right-boundary bias magnitude relationship is directly derived as: 
\begin{multline*}
0 = \textcolor{biascolor}{\text{Bias}_\text{DGPO}^\text{HN}} = \text{Bias}_\text{GRPO}^\text{HN} = \text{Bias}_\text{GPPO}^\text{HN} \\
= \text{Bias}_\text{CE}^\text{HN} < \text{Bias}_\text{CISPO}^\text{HN} = \text{Bias}_\text{ASPO}^\text{HN}.
\end{multline*}

\subsection{Derivation of Learning Rate Scaling}
\label{app:lr_scaling}

To ensure consistent training dynamics across models of varying sizes (1.5B, 7B, and 14B), we employ a learning rate scaling rule based on the principle of \textit{Constant Total Gradient Variance}. Below, we provide the formal derivation of this scaling law.

\paragraph{Assumptions.}
Consider two models with identical architecture but different parameter counts, denoted by $N$. We make the following standard assumptions for large-scale model training:
\begin{itemize}
    \item The training data, batch size, and optimizer configuration remain constant.
    \item The gradient variance of a single parameter, denoted as $\sigma^2$, is constant and independent of the total model size (assuming high redundancy in LLM parameters).
    \item The gradients of individual parameters are approximately independent.
\end{itemize}

\begin{table*}[htbp]
    \centering
    \small
    \caption{Reverse Left-Boundary bias among various policy optimization algorithms}
    \label{tab:LP_bias}
    \begin{tabular}{l l}
        \toprule
        \textbf{Algorithm} & \textbf{Reverse Left-Boundary Bias Magnitude}\\
        \midrule
        GRPO, GPPO & \multirow{2}{*}{$\begin{aligned}
            \text{Bias}_\text{X}^\text{LP} = \left\| \nabla_\theta\mathcal{J}_{\text{X}}^\text{LP}(\theta) - \nabla_\theta\mathcal{J}_{\text{PG}}^\text{LP}(\theta) \right\| = 0
        \end{aligned}$} \\
        CE-GPPO, DGPO & \\
        \addlinespace[4pt]
        \multirow{1}{*}{CISPO, ASPO} &
        $\begin{aligned}[t]
            \text{Bias}_\text{CISPO}^\text{LP} &= \left\| \nabla_\theta\mathcal{J}_{\text{CISPO}}^\text{LP}(\theta) - \nabla_\theta\mathcal{J}_{\text{PG}}^\text{LP}(\theta) \right\| \\
            &= \left\| \mathbb{E}_{q \sim \mathcal{D},\, o \sim \pi_{\theta_{\text{old}}}} \left[ v_t^{\text{LP}} \cdot (1-\varepsilon_\text{low}-r_t) \cdot A_t \nabla_\theta \log \pi_\theta(o_t | q, o_{<t}) \right] \right\| \neq 0^*
        \end{aligned}$ \\
        \bottomrule
    \end{tabular}

    \vspace{0.2cm}

    \centering
    \small
    \caption{Reverse Right-Boundary bias among various policy optimization algorithms}
    \label{tab:HN_bias}
    \begin{tabular}{l l}
        \toprule
        \textbf{Algorithm} & \textbf{Reverse Left-Boundary Bias Magnitude}\\
        \midrule
        GRPO, GPPO & \multirow{2}{*}{$\begin{aligned}
            \text{Bias}_\text{X}^\text{HN} = \left\| \nabla_\theta\mathcal{J}_{\text{X}}^\text{HN}(\theta) - \nabla_\theta\mathcal{J}_{\text{PG}}^\text{HN}(\theta) \right\| = 0
        \end{aligned}$} \\
        CE-GPPO, DGPO & \\
        \addlinespace[4pt]
        \multirow{1}{*}{CISPO, ASPO} &
        $\begin{aligned}[t]
            \text{Bias}_\text{CISPO}^\text{HN} &= \left\| \nabla_\theta\mathcal{J}_{\text{CISPO}}^\text{HN}(\theta) - \nabla_\theta\mathcal{J}_{\text{PG}}^\text{HN}(\theta) \right\| \\
            &= \left\| \mathbb{E}_{q \sim \mathcal{D},\, o \sim \pi_{\theta_{\text{old}}}} \left[ v_t^{\text{HN}} \cdot [r_t-(1+\varepsilon_\text{high})] \cdot A_t \nabla_\theta \log \pi_\theta(o_t | q, o_{<t}) \right] \right\| \neq 0^*
        \end{aligned}$ \\
        \bottomrule
    \end{tabular}
\end{table*}

\begin{figure*}[t]
\begin{equation} \label{eq:RLB_gradients}
\begin{cases}
\nabla_\theta\mathcal{J}_{\text{PG}}^\text{LP}(\theta) = \mathbb{E}_{q \sim \mathcal{D},\, o \sim \pi_{\theta_{\text{old}}}} \left[ v_t^{\text{LP}} \cdot r_t \cdot A_t \nabla_\theta \log \pi_\theta(o_t | q, o_{<t})\right] \\
\nabla_\theta\mathcal{J}_{\text{GRPO}}^\text{LP}(\theta) = \mathbb{E}_{q \sim \mathcal{D},\, o \sim \pi_{\theta_{\text{old}}}} \left[ v_t^{\text{LP}}  \cdot r_t \cdot  A_t \nabla_\theta \log \pi_\theta(o_t | q, o_{<t}) \right] \\
\nabla_\theta\mathcal{J}_{\text{CISPO}}^\text{LP}(\theta) = \mathbb{E}_{q \sim \mathcal{D},\, o \sim \pi_{\theta_{\text{old}}}} \left[ v_t^{\text{LP}} \cdot (1-\varepsilon_\text{low}) \cdot A_t \nabla_\theta \log \pi_\theta(o_t | q, o_{<t}) \right] \\ 
\nabla_\theta\mathcal{J}_{\text{GPPO}}^\text{LP}(\theta) = \mathbb{E}_{q \sim \mathcal{D},\, o \sim \pi_{\theta_{\text{old}}}} \left[ v_t^{\text{LP}} \cdot r_t \cdot A_t \nabla_\theta \log \pi_\theta(o_t | q, o_{<t}) \right] \\
\nabla_\theta\mathcal{J}_{\text{CE}}^\text{LP} (\theta) = \mathbb{E}_{q \sim \mathcal{D},\, o \sim \pi_{\theta_{\text{old}}}} \left[ v_t^{\text{LP}} \cdot r_t \cdot A_t \nabla_\theta \log \pi_\theta(o_t | q, o_{<t}) \right] \\
\nabla_\theta\mathcal{J}_{\text{ASPO}}^\text{LP} (\theta) = \mathbb{E}_{q \sim \mathcal{D},\, o \sim \pi_{\theta_{\text{old}}}} \left[  v_t^{\text{LP}} \cdot (1-\varepsilon_\text{low}) \cdot  A_t \nabla_\theta \log \pi_\theta(o_t | q, o_{<t}) \right] \\
\nabla_\theta\mathcal{J}_{\text{DGPO}}^\text{LP} (\theta) = \mathbb{E}_{q \sim \mathcal{D},\, o \sim \pi_{\theta_{\text{old}}}} \left[  v_t^{\text{LP}} \cdot r_t \cdot A_t \nabla_\theta \log \pi_\theta(o_t | q, o_{<t}) \right]
\end{cases}
\end{equation}

\vspace{0.2cm}

\begin{equation} \label{eq:RRB_gradients}
\begin{cases}
\nabla_\theta\mathcal{J}_{\text{PG}}^\text{HN}(\theta) = \mathbb{E}_{q \sim \mathcal{D},\, o \sim \pi_{\theta_{\text{old}}}} \left[ v_t^{\text{HN}} \cdot r_t \cdot A_t \nabla_\theta \log \pi_\theta(o_t | q, o_{<t})\right] \\
\nabla_\theta\mathcal{J}_{\text{GRPO}}^\text{HN}(\theta) = \mathbb{E}_{q \sim \mathcal{D},\, o \sim \pi_{\theta_{\text{old}}}} \left[ v_t^{\text{HN}}  \cdot r_t \cdot  A_t \nabla_\theta \log \pi_\theta(o_t | q, o_{<t}) \right] \\
\nabla_\theta\mathcal{J}_{\text{CISPO}}^\text{HN}(\theta) = \mathbb{E}_{q \sim \mathcal{D},\, o \sim \pi_{\theta_{\text{old}}}} \left[ v_t^{\text{HN}} \cdot (1+\varepsilon_\text{high}) \cdot A_t \nabla_\theta \log \pi_\theta(o_t | q, o_{<t}) \right] \\ 
\nabla_\theta\mathcal{J}_{\text{GPPO}}^\text{HN}(\theta) = \mathbb{E}_{q \sim \mathcal{D},\, o \sim \pi_{\theta_{\text{old}}}} \left[ v_t^{\text{HN}} \cdot r_t \cdot A_t \nabla_\theta \log \pi_\theta(o_t | q, o_{<t}) \right] \\
\nabla_\theta\mathcal{J}_{\text{CE}}^\text{HN} (\theta) = \mathbb{E}_{q \sim \mathcal{D},\, o \sim \pi_{\theta_{\text{old}}}} \left[ v_t^{\text{HN}} \cdot r_t \cdot A_t \nabla_\theta \log \pi_\theta(o_t | q, o_{<t}) \right] \\
\nabla_\theta\mathcal{J}_{\text{ASPO}}^\text{HN} (\theta) = \mathbb{E}_{q \sim \mathcal{D},\, o \sim \pi_{\theta_{\text{old}}}} \left[  v_t^{\text{HN}} \cdot (1+\varepsilon_\text{high}) \cdot  A_t \nabla_\theta \log \pi_\theta(o_t | q, o_{<t}) \right] \\
\nabla_\theta\mathcal{J}_{\text{DGPO}}^\text{HN} (\theta) = \mathbb{E}_{q \sim \mathcal{D},\, o \sim \pi_{\theta_{\text{old}}}} \left[  v_t^{\text{HN}} \cdot r_t \cdot A_t \nabla_\theta \log \pi_\theta(o_t | q, o_{<t}) \right]
\end{cases}
\end{equation}
\end{figure*}

\paragraph{Total Gradient Variance.}
Let $\nabla_\theta \mathcal{L}$ represent the gradient vector of the loss function with respect to the model parameters $\theta \in \mathbb{R}^N$. Based on the independence assumption, the variance of the total gradient norm is the sum of the variances of individual parameter gradients:
\begin{equation}
    \text{Var}(\nabla_\theta \mathcal{L}) = \sum_{i=1}^N \text{Var}(g_i) = N \cdot \sigma^2
\end{equation}

\paragraph{Stability Condition.}
To maintain a consistent convergence speed across different scales, we require the variance of the parameter update step (the ``step size'' in the parameter space) to remain constant. Let $\eta$ be the learning rate. The update step is $\Delta \theta = \eta \cdot \nabla_\theta \mathcal{L}$. The variance of this update step is:
\begin{align}
    \text{Var}(\Delta \theta) &= \text{Var}(\eta \cdot \nabla_\theta \mathcal{L}) & \nonumber \\
    &= \eta^2 \cdot \text{Var}(\nabla_\theta \mathcal{L}) = \eta^2 N \sigma^2
\end{align}
We define a stability constant $C$ such that:
\begin{equation}
    \label{eq:stability}
    \eta^2 N \sigma^2 = C
\end{equation}

\paragraph{Scaling Law Derivation.}
Let $(\eta_1, N_1)$ be the configuration for the base model (1.5B) and $(\eta_2, N_2)$ be the configuration for the target model (e.g., 7B or 14B). From Eq. \eqref{eq:stability}, we have:
\begin{equation}
    \eta_1^2 N_1 \sigma^2 = \eta_2^2 N_2 \sigma^2 = C
\end{equation}
Eliminating constant terms $\sigma^2$ and $C$, we obtain the relationship:
\begin{equation}
    \eta_1^2 N_1 = \eta_2^2 N_2 \implies \eta_2 = \eta_1 \sqrt{\frac{N_1}{N_2}}
\end{equation}
This establishes the \textbf{Inverse Square Root Scaling Law} relative to the number of parameters.

\paragraph{Numerical Verification.}
Applying this rule to our experimental settings with the base learning rate $\eta_{1.5B} = 1.0 \times 10^{-6}$:
\begin{itemize}
    \item \textbf{For 7B Model:}
    \begin{flalign*}
        \eta_{7B} &= 1.0 \times 10^{-6} \times \sqrt{\frac{1.5}{7}} & \nonumber \\
        &\approx 1.0 \times 10^{-6} \times 0.4629 \approx 4.63 \times 10^{-7}
    \end{flalign*}
    \item \textbf{For 14B Model:}
    \begin{flalign*}
        \eta_{14B} &= 1.0 \times 10^{-6} \times \sqrt{\frac{1.5}{14}} & \nonumber \\
        &\approx 1.0 \times 10^{-6} \times 0.3273 \approx 3.27 \times 10^{-7}
    \end{flalign*}
\end{itemize}
These calculated values correspond exactly to the learning rates reported in Table \ref{tab:detailed_hyperparameters}.

\section{Implementation Details}
\label{app:implementation_details}

\subsection{Infrastructure and Environment}
\paragraph{Computational Resources.}
Our experiments are conducted on a high-performance computing cluster consisting of 30 nodes. Each node is equipped with 8 $\times$ NVIDIA A100 GPUs (80GB VRAM), interconnected via NVLink for high-bandwidth intra-node communication. The multi-node training relies on a robust Ethernet/InfiniBand fabric to ensure synchronization efficiency. The total effective training time for the largest model (14B) was approximately 500 hours.

\paragraph{Software Stack.}
We build our reinforcement learning pipeline upon VeRL ~\cite{sheng2025hybridflow}, a flexible framework designed for post-training.
\begin{itemize}
    \item \textbf{Training Backend:} We utilize Fully Sharded Data Parallel (FSDP) for distributed training. This setup manages memory efficiency through parameter sharding and offloading, allowing us to train 14B models with full-parameter updates without memory overflow.
    \item \textbf{Inference Engine:} To maximize rollout throughput, we integrate vLLM as the inference backend. We leverage its \textbf{PagedAttention} mechanism to efficiently manage Key-Value (KV) cache memory, significantly reducing fragmentation during the generation of long reasoning chains. We align the environment configurations (CUDA version 12.4, PyTorch version 2.6.0) across all nodes to prevent numerical discrepancies.
\end{itemize}

\subsection{Data and Evaluation Benchmarks}
\paragraph{Training Dataset.}
We utilize the DAPO-Math-17k dataset for training. We employ the official tokenizer corresponding to the Qwen2.5-Math series to ensure consistent token mapping between the pre-trained backbone and the RL fine-tuning stage. The maximum response length is set to 8192 tokens. This length is empirically chosen to sufficiently accommodate the chain-of-thought reasoning steps required by benchmarks like AIME and MATH, while maintaining high training throughput.

\paragraph{Benchmark Details} To rigorously evaluate the mathematical reasoning capabilities of our models, we selected a suite of diverse benchmarks. These datasets cover a spectrum of difficulty levels, from foundational high school mathematics to expert-level olympiad problems. Table \ref{tab:benchmark_details} provides a comprehensive overview of their characteristics.

\begin{table*}[h]
\centering
\small
\renewcommand{\arraystretch}{1.5}
\renewcommand\tabularxcolumn[1]{m{#1}}
\begin{tabularx}{\textwidth}{
    l
    >{\hsize=0.8\hsize\linewidth=\hsize\arraybackslash}X 
    >{\hsize=1.2\hsize\linewidth=\hsize\raggedright\arraybackslash}X
}
\toprule
\textbf{Dataset} & \textbf{Core Description} & \textbf{Key Characteristics} \\
\midrule

\textbf{AIME 2024} & 
The 2024 edition of the American Invitational Mathematics Examination, serving as a bridge between the AMC and the USAMO. & 
\begin{itemize}[leftmargin=*, nosep, after=\vspace{0pt}]
    \item Focuses on arithmetic precision and number theory.
    \item Requires answers in a strict integer format (000--999).
    \item Tests robustness against recent contamination.
\end{itemize} \\
\midrule

\textbf{AIME 2025} & 
The most recent iteration of the AIME competition, representing a strictly ``held-out'' set for evaluating generalization to unseen problems. & 
\begin{itemize}[leftmargin=*, nosep, after=\vspace{0pt}]
    \item Zero data contamination risk due to recency.
    \item High complexity requiring multi-step logical chains.
    \item Validates the model's potential for future reasoning.
\end{itemize} \\
\midrule

\textbf{AMC 2023} & 
Selected problems from the 2023 American Mathematics Competitions (AMC 10/12), representing the entry-level olympiad difficulty. & 
\begin{itemize}[leftmargin=*, nosep, after=\vspace{0pt}]
    \item Covers broad topics: Algebra, Geometry, Counting.
    \item Functions as a baseline for competitive math ability.
    \item Requires mapping multiple-choice logic to open-ended generation.
\end{itemize} \\
\midrule

\textbf{MATH-500} & 
A curated subset of 500 representative problems from the widely used MATH dataset, designed by OpenAI for efficient evaluation. & 
\begin{itemize}[leftmargin=*, nosep, after=\vspace{0pt}]
    \item Spans 7 categories including Calculus and Probability.
    \item Reduces evaluation costs while maintaining distribution fidelity.
    \item Heavily relies on LaTeX understanding.
\end{itemize} \\
\midrule

\textbf{Minerva} & 
A collection of technical mathematics problems derived from scientific papers and undergraduate-level coursework. & 
\begin{itemize}[leftmargin=*, nosep, after=\vspace{0pt}]
    \item Involves higher-order symbolic reasoning.
    \item Contains domain-specific vocabulary and notation.
    \item Tests capabilities beyond standard competition math.
\end{itemize} \\
\midrule

\textbf{OlympiadBench} & 
A comprehensive aggregate of international mathematics competitions (e.g., IMO, CMO) spanning multiple languages and formats. & 
\begin{itemize}[leftmargin=*, nosep, after=\vspace{0pt}]
    \item Represents the upper bound of mathematical reasoning.
    \item Includes theorem proving and fill-in-the-blank types.
    \item Challenges the model's cross-lingual mathematical logic.
\end{itemize} \\

\bottomrule
\end{tabularx}
\caption{Detailed comparison of the mathematical reasoning benchmarks used in this study.}
\label{tab:benchmark_details}
\end{table*}

The datasets used for evaluation can be accessed via the following repositories:
\begin{itemize}[leftmargin=1cm]
    \item \textbf{AIME 2024:} \url{https://huggingface.co/datasets/math-ai/aime24}
    \item \textbf{AIME 2025:} \url{https://huggingface.co/datasets/math-ai/aime25}
    \item \textbf{AMC 2023:} \url{https://huggingface.co/datasets/math-ai/amc23}
    \item \textbf{MATH-500:} \url{https://huggingface.co/datasets/HuggingFaceH4/MATH-500}
    \item \textbf{Minerva:} \url{https://huggingface.co/datasets/math-ai/minervamath}
    \item \textbf{OlympiadBench:} \url{https://huggingface.co/datasets/math-ai/olympiadbench}
\end{itemize}

\paragraph{Reward Function.}
We employ a strict rule-based reward mechanism to rigorously verify the correctness of the generated solutions.
\begin{itemize}
    \item \textbf{Format Verification:} We first check if the output follows the required format (e.g., enclosing the answer in \texttt{\textbackslash boxed\{\}}).
    \item \textbf{Correctness Check:} Using the \texttt{math\_verify} toolkit, we compare the extracted answer against the ground truth. A reward of $r=1$ is assigned for a correct match, and $r=-1$ otherwise. This binary reward setting poses a significant challenge for exploration, as no partial rewards are applied.
\end{itemize}

\subsection{Training Protocols}
\paragraph{Optimization Strategy.}
We employ the AdamW optimizer with $\beta_1 = 0.9$, $\beta_2 = 0.95$, and a small weight decay. Unlike standard pre-training schedules, we adopt a \textbf{constant learning rate} strategy without warm-up or cosine decay. This design choice eliminates the confounding factors of learning rate scheduling, allowing us to attribute performance gains solely to the algorithmic improvements. To ensure consistent convergence dynamics across different model scales (1.5B, 7B, 14B), we calibrate the learning rate based on the \textit{Constant Total Gradient Variance} principle.

\paragraph{Numerical Precision and Stability.}
All models are trained using \texttt{bfloat16} precision. We apply gradient clipping with a norm threshold of 1.0 to mitigate gradient explosion. Random seeds for model initialization, data shuffling, and sampling are fixed to 42 to ensure reproducibility.

\subsection{Hyperparameter Specifications}
Table \ref{tab:detailed_hyperparameters} summarizes the key hyperparameters used across all experiments.

\begin{table*}[t!]
\centering
\scriptsize
\renewcommand{\arraystretch}{1.0}
\caption{Detailed hyperparameter configurations for all experiments. To ensure a fair comparison, we maintain identical training settings across all algorithms, varying only the learning rate according to the model scale.}
\resizebox{\textwidth}{!}{%
\begin{tabular}{llcl}
\toprule
\textbf{Model Scale} & \textbf{Algorithm} & \textbf{Learning Rate} & \textbf{Specific Hyperparameters} \\
\midrule
\multicolumn{4}{c}{\textit{Common Settings: Mini-batch Size = 32, Rollout Batch Size = 512, Max Length = 8192}} \\
\midrule

\multirow{6}{*}{DeepSeek-R1-Distill-Qwen-1.5B} 
 & GRPO & \multirow{6}{*}{$1.0 \times 10^{-6}$} & $\varepsilon_\text{low} = \varepsilon_\text{high} = 0.2$ \\
 & CISPO &  & $\varepsilon_\text{low} = \varepsilon_\text{high} = 0.2$ \\
 & GPPO &  & $\varepsilon_\text{low} = \varepsilon_\text{high} = 0.2$ \\
 & CE-GPPO &  & $\varepsilon_\text{low} = \varepsilon_\text{high} = 0.2, \beta_1 = 0.75, \beta_2 = 1$ \\
 & ASPO &  & $\varepsilon_\text{low} = \varepsilon_\text{high} = 0.2, \varepsilon'_\text{low}=0.33, \varepsilon'_\text{high}=3$ \\
 & \textbf{DGPO} &  & $\varepsilon_\text{low} = \varepsilon_\text{high} = 0.2, n=2, m=2$ \\
\midrule

\multirow{6}{*}{DeepSeek-R1-Distill-Qwen-7B} 
 & GRPO & \multirow{6}{*}{$4.63 \times 10^{-7}$} & $\varepsilon_\text{low} = \varepsilon_\text{high} = 0.2$ \\
 & CISPO &  & $\varepsilon_\text{low} = \varepsilon_\text{high} = 0.2$ \\
 & GPPO &  & $\varepsilon_\text{low} = \varepsilon_\text{high} = 0.2$ \\
 & CE-GPPO &  & $\varepsilon_\text{low} = \varepsilon_\text{high} = 0.2, \beta_1 = 0.75, \beta_2 = 1$ \\
 & ASPO &  & $\varepsilon_\text{low} = \varepsilon_\text{high} = 0.2, \varepsilon'_\text{low}=0.33, \varepsilon'_\text{high}=3$ \\
 & \textbf{DGPO} &  & $\varepsilon_\text{low} = \varepsilon_\text{high} = 0.2, n=1, m=2$ \\
\midrule

\multirow{6}{*}{DeepSeek-R1-Distill-Qwen-14B} 
 & GRPO & \multirow{6}{*}{$3.27 \times 10^{-7}$} & $\varepsilon_\text{low} = \varepsilon_\text{high} = 0.2$ \\
 & CISPO &  & $\varepsilon_\text{low} = \varepsilon_\text{high} = 0.2$ \\
 & GPPO &  & $\varepsilon_\text{low} = \varepsilon_\text{high} = 0.2$ \\
 & CE-GPPO &  & $\varepsilon_\text{low} = \varepsilon_\text{high} = 0.2, \beta_1 = 0.75, \beta_2 = 1$ \\
 & ASPO &  & $\varepsilon_\text{low} = \varepsilon_\text{high} = 0.2, \varepsilon'_\text{low}=0.33, \varepsilon'_\text{high}=3$ \\
 & \textbf{DGPO} &  & $\varepsilon_\text{low} = \varepsilon_\text{high} = 0.2, n=1, m=2$ \\
\bottomrule
\end{tabular}%
}
\label{tab:detailed_hyperparameters}
\end{table*}

\begin{figure*}[t!]
    \centering
    \begin{subfigure}{0.32\textwidth}
        \centering
        \includegraphics[width=\linewidth]{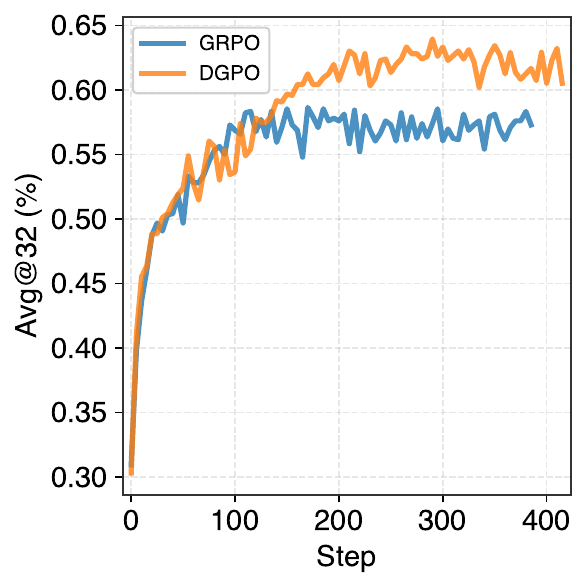} 
        \caption{AIME 2024 (Avg@32)}
    \end{subfigure}
    \hfill
    \begin{subfigure}{0.32\textwidth}
        \centering
        \includegraphics[width=\linewidth]{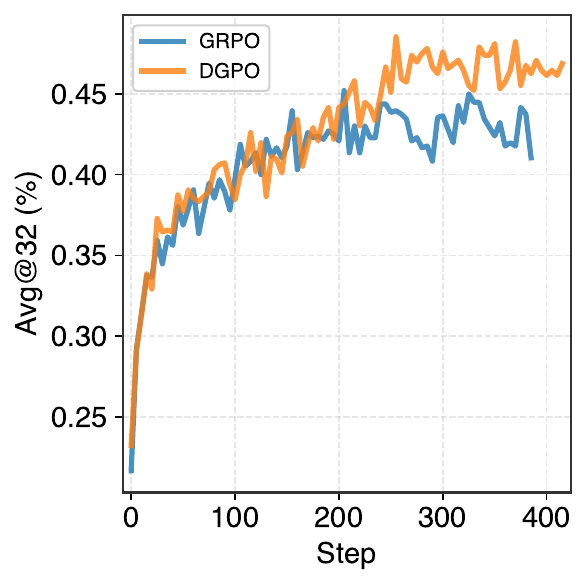}
        \caption{AIME 2025 (Avg@32)}
    \end{subfigure}
    \hfill
    \begin{subfigure}{0.32\textwidth}
        \centering
        \includegraphics[width=\linewidth]{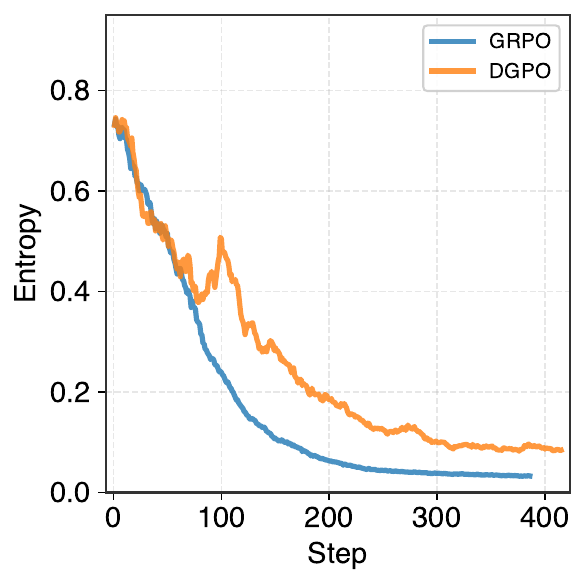}
        \caption{Policy Entropy}
    \end{subfigure}
    \caption{Training dynamics of DeepSeek-R1-Distill-Qwen-14B comparing GRPO and DGPO.}
    \label{fig:14b_dynamics}
\end{figure*}

\paragraph{Controlled Variables.}
To ensure a fair comparison and strictly isolate the impact of different gradient weighting strategies, we unify the clipping thresholds across all experiments. Specifically, for all methods involving trust region clipping or boundary definitions (including GRPO, CISPO, GPPO, CE-GPPO, ASPO and our DGPO), we fix:
\begin{equation*}
    \varepsilon_\text{low} = \varepsilon_\text{high} = 0.2
\end{equation*}
This rigorous control ensures that any observed performance differences are  primarily attributable to the gradient dynamics rather than minor variations in the trust region size.

\paragraph{Common Configurations.}
We set the KL coefficient $\beta_\text{KL} = 0$. By removing the explicit KL penalty, we rely implicitly on the trust region constraints imposed by the clipping mechanisms to prevent policy collapse. The rollout generation uses a temperature of 1.0 and a top-$p$ of 1.0.

\paragraph{Method-Specific Settings.}
The method-specific configurations are as follows:
\begin{itemize}
    \item \textbf{CE-GPPO:} Adopts scaled soft clipping with $\beta_1 = 0.75$ and $\beta_2 = 1.0$, as recommended in the original paper.
    \item \textbf{ASPO:} Use $\varepsilon'_\text{low}=0.33$ and $\varepsilon'_\text{high}=3$ as soft dual clip threshold.
    \item \textbf{DGPO:} We introduce the decoupled decay parameters $n$ and $m$:
    \begin{itemize}
        \item \textbf{1.5B:} $n=2$, $m=2$.
        \item \textbf{7B \& 14B:} $n=1$, $m=2$.
    \end{itemize}
    The continuity constants $C_\text{left}$ and $C_\text{right}$ are automatically calculated based on $\pi_{\theta_\text{old}}$.
\end{itemize}

\section{Additional Experimental Results}
\label{app:additional_results}

\subsection{14B Model Training Dynamics}
\label{app:14b_dynamics}

We visualize the training trajectories of the DeepSeek-R1-Distill-Qwen-14B model in Figure \ref{fig:14b_dynamics}. Similar to the 1.5B and 7B models, DGPO on 14B scale demonstrates faster convergence and higher asymptotic performance on AIME benchmarks while maintaining a stable entropy reduction curve, avoiding the collapse issues seen in RLVR.

\subsection{Detailed 14B Performance Comparison}
\label{app:14b_full_table}

Table \ref{tab:14b_full_results} provides comprehensive evaluation results for 14B model, including both Avg@32 (expected performance) and Pass@32 (potential capability).

\subsection{Pass@K Statistics on AIME Benchmarks} 
\label{app:detailed_passk}

We report Pass@K ($k \in \{1, 2, 4, 8, 16, 32\}$) metrics specifically for the \textbf{AIME 2024} and \textbf{AIME 2025} benchmarks across 1.5B, 7B, and 14B scales (Table \ref{tab:full_passk_aime}). The results demonstrate that DGPO consistently achieves higher coverage of the solution space (higher Pass@K) compared to baselines, particularly as $k$ increases.

\definecolor{lavender}{RGB}{240, 240, 245}

\begin{table*}[h]
    \centering
    \small
    \setlength{\tabcolsep}{3.3pt}
    \caption{Full performance comparison on DeepSeek-R1-Distill-Qwen-14B.}
    \begin{tabular}{lcccccccccccccc}
    \toprule
    \multirow{2}{*}{\textbf{Method}} & \multicolumn{2}{c}{\textbf{AIME24}} & \multicolumn{2}{c}{\textbf{AIME25}} & \multicolumn{2}{c}{\textbf{AMC23}} & \multicolumn{2}{c}{\textbf{MATH500}} & \multicolumn{2}{c}{\textbf{Minerva}} & \multicolumn{2}{c}{\textbf{Olympiad}} & \multicolumn{2}{c}{\textbf{Avg.}} \\
    \cmidrule(lr){2-3} \cmidrule(lr){4-5} \cmidrule(lr){6-7} \cmidrule(lr){8-9} \cmidrule(lr){10-11} \cmidrule(lr){12-13} \cmidrule(lr){14-15}
    & A@32 & P@32 & A@32 & P@32 & A@32 & P@32 & A@32 & P@32 & A@32 & P@32 & A@32 & P@32 & A@32 & P@32 \\
    \midrule
    \modelheader{lavender}{DeepSeek-R1-Distill-Qwen-14B} \\ 
    \midrule
    GRPO & 56.6 & 82.9 & 40.5 & \best{67.8} & 92.2 & 97.4 & 66.5 & 70.4 & 22.6 & 35.3 & 43.2 & 50.7 & 53.6 & 67.4 \\
    \rowcolor{ourshighlight}
    \textbf{DGPO} & \best{63.3} & \best{86.2} & \best{47.6} & 66.4 & \best{93.9} & \best{99.1} & \best{67.0} & \best{75.1} & \best{23.0} & \best{37.9} & \best{45.1} & \best{57.4} & \best{56.7} & \best{70.4} \\
    \bottomrule
    \end{tabular}
    \label{tab:14b_full_results}

    \vspace{0.2cm}

    \centering
    \tiny
    \renewcommand{\arraystretch}{1}
    \setlength{\tabcolsep}{12pt}
    \caption{Detailed Pass@K performance on AIME 2024 and AIME 2025 across all model scales.}
    \resizebox{\textwidth}{!}{
    \begin{tabular}{lcccccc}
    \toprule
    \textbf{Method} & \textbf{P@1} & \textbf{P@2} & \textbf{P@4} & \textbf{P@8} & \textbf{P@16} & \textbf{P@32} \\
    \midrule
    
    \multicolumn{7}{c}{\cellcolor{mistyblue}\textbf{DeepSeek-R1-Distill-Qwen-1.5B}} \\
    \midrule
    \multicolumn{7}{c}{\textit{\textbf{Dataset: AIME 2024}}} \\
    GRPO    & 33.2 & 43.2 & 53.2 & 61.6 & 67.9 & 71.8 \\
    CISPO   & 34.8 & 44.1 & 52.0 & 58.7 & 64.4 & 69.1 \\
    GPPO    & 29.6 & 38.4 & 47.1 & 53.6 & 57.9 & 60.5 \\
    CE-GPPO & 35.1 & 44.4 & 52.6 & 60.0 & 66.2 & 70.2 \\
    ASPO    & 36.4 & 44.4 & 53.1 & 61.7 & 68.5 & 73.2 \\
    \rowcolor{ourshighlight}
    \textbf{DGPO} & \best{43.3} & \best{53.4} & \best{63.1} & \best{70.6} & \best{75.6} & \best{79.3} \\
    \addlinespace[4pt]
    \multicolumn{7}{c}{\textit{\textbf{Dataset: AIME 2025}}} \\
    GRPO    & 27.7 & 33.0 & 38.1 & 42.6 & 46.5 & 49.9 \\
    CISPO   & 25.8 & 31.8 & 37.6 & 43.6 & 49.2 & 53.3 \\
    GPPO    & 23.5 & 29.3 & 34.3 & 40.0 & 46.4 & 51.9 \\
    CE-GPPO & 27.7 & 33.1 & 38.3 & 44.1 & 50.1 & 55.1 \\
    ASPO    & 28.3 & 33.0 & 37.1 & 41.2 & 45.9 & 51.5 \\
    \rowcolor{ourshighlight}
    \textbf{DGPO} & \best{32.8} & \best{38.3} & \best{43.6} & \best{48.6} & \best{52.7} & \best{56.1} \\
    
    \midrule
    \multicolumn{7}{c}{\cellcolor{almond}\textbf{DeepSeek-R1-Distill-Qwen-7B}} \\
    \midrule
    \multicolumn{7}{c}{\textit{\textbf{Dataset: AIME 2024}}} \\
    GRPO    & 48.2  & 58.0  & 66.6  & 73.4  & 78.6  & \best{82.5} \\
    CISPO   & 51.6  & 60.9  & 68.5  & 73.5  & 75.8  & 76.6 \\
    GPPO    & 43.1  & 52.1  & 59.3  & 65.0  & 69.7  & 72.5 \\
    CE-GPPO & 48.7  & 55.7  & 62.3  & 68.1  & 73.0  & 76.9 \\
    ASPO    & 51.8  & 61.2  & 68.5  & 73.4  & 76.7 & 79.6 \\
    \rowcolor{ourshighlight}
    \textbf{DGPO} & \best{55.5} & \best{64.1} & \best{71.6}  & \best{77.0} & \best{80.2} & 81.9 \\
    \addlinespace[4pt]
    \multicolumn{7}{c}{\textit{\textbf{Dataset: AIME 2025}}} \\
    GRPO    & 37.4  & 44.4  & 50.8  & 55.2  & 58.1  & 60.5 \\
    CISPO   & 38.2  & 45.2  & 51.7  & 56.5  & 60.8 & 65.4 \\
    GPPO    & 31.7  & 38.5  & 45.8  & 53.0  & 58.6  & 62.5 \\
    CE-GPPO & 36.4  & 43.1  & 49.1  & 53.9  & 57.4  & 60.4 \\
    ASPO    & 37.1  & 43.3  & 48.4  & 51.2  & 52.6  & 54.1 \\
    \rowcolor{ourshighlight}
    \textbf{DGPO} & \best{43.1} & \best{49.7} & \best{55.2} & \best{60.1} & \best{64.6} & \best{68.0} \\
    
    \midrule
    \multicolumn{7}{c}{\cellcolor{lavender}\textbf{DeepSeek-R1-Distill-Qwen-14B}} \\
    \midrule
    \multicolumn{7}{c}{\textit{\textbf{Dataset: AIME 2024}}} \\
    GRPO    & 56.6 & 69.3 & 76.0 & 79.8 & 81.9 & 82.9 \\
    \rowcolor{ourshighlight}
    \textbf{DGPO} & \best{63.3} & \best{70.4} & \best{76.4} & \best{81.1} & \best{84.6} & \best{86.2} \\
    \addlinespace[4pt]
    \multicolumn{7}{c}{\textit{\textbf{Dataset: AIME 2025}}} \\
    GRPO    & 40.5 & 48.4 & 54.2 & 60.1 & \best{64.7} & \best{67.8} \\
    \rowcolor{ourshighlight}
    \textbf{DGPO} & \best{47.6} & \best{53.7} & \best{58.7} & \best{61.7} & 63.9 & 66.4 \\
    
    \bottomrule
    \end{tabular}
    }
    \label{tab:full_passk_aime}
\end{table*}

\end{document}